%% file: iclr2023_conference.tex
\documentclass{article} 
\usepackage{iclr2023_conference,times}

\input{math_commands.tex}

\usepackage{hyperref}
\usepackage{url}
\usepackage{graphicx}
\usepackage{booktabs}
\usepackage{graphicx}
\usepackage{bm}
\usepackage{amsmath}
\usepackage{amsfonts}
\usepackage{adjustbox}
\usepackage{amsthm}
\usepackage{cancel}
\usepackage{xcolor}
\usepackage[linesnumbered,ruled]{algorithm2e}
\usepackage{etoolbox}
\usepackage{caption}
\usepackage{amssymb}
\makeatletter
\patchcmd{\@algocf@start}
  {-1.5em}
  {0pt}
  {}{}
\makeatother

\newtheorem{theorem}{Theorem}
\newcommand\bmt[1]{\bm{\tilde{#1}}}
\newtheorem{innercustomthm}{Theorem}
\newenvironment{customthm}[1]
  {\renewcommand\theinnercustomthm{#1}\innercustomthm}
  {\endinnercustomthm}

\title{Correcting the Sub-optimal Bit Allocation}

\author{Tongda Xu\textsuperscript{\rm 1}, Han Gao\textsuperscript{\rm 2,3}, Yuanyuan Wang\textsuperscript{\rm 2}, Hongwei Qin\textsuperscript{\rm 2}, Yan Wang \textsuperscript{\rm 1}\thanks{To whom the correspondence should be addressed.},\\ \textbf{Jingjing Liu\textsuperscript{\rm 1,4}, Ya-Qin Zhang\textsuperscript{\rm 1,4,5}}\\
\textsuperscript{\rm 1}Institute for AI Industry Research (AIR), Tsinghua University,\\ \textsuperscript{\rm 2}SenseTime Research,\textsuperscript{\rm 3}University of Electronic Science and Technology of China,\\
\textsuperscript{\rm 4}School of Vehicle and Mobility, Tsinghua University,\\
\textsuperscript{\rm 5}Department of Computer Science and Technology, Tsinghua University\\
\texttt{\{xutongda,wangyan,JJLiu,zhangyaqin\}\@air.tsinghua.edu.cn}\\
\texttt{\{gaohan1,wangyuanyuan,qinhongwei\}\@sensetime.com}
}

\iclrfinalcopy 
\begin{document}

\maketitle

\begin{abstract}
    In this paper, we investigate the problem of bit allocation in Neural Video Compression (NVC). First, we reveal that a recent bit allocation approach claimed to be optimal is, in fact, sub-optimal due to its implementation. Specifically, we find that its sub-optimality lies in the improper application of semi-amortized variational inference (SAVI) on latent with non-factorized variational posterior. Then, we show that the corrected version of SAVI on non-factorized latent requires recursively applying back-propagating through gradient ascent, based on which we derive the corrected optimal bit allocation algorithm. Due to the computational in-feasibility of the corrected bit allocation, we design an efficient approximation to make it practical. Empirical results show that our proposed correction significantly improves the incorrect bit allocation in terms of R-D performance and bitrate error, and outperforms all other bit allocation methods by a large margin. The source code is provided in the supplementary material.
\end{abstract}

\section{Introduction}

Recently, bit allocation for Neural Video Compression (NVC) has drawn growing attention thanks to its great potential in boosting compression performance. Due to the frame reference structure in video coding, it is sub-optimal to use the same R-D (Rate-Distortion) trade-off parameter $\lambda$ for all frames. In bit allocation task, bitrate is allocated  to different frames/regions to minimize R-D cost $R+\lambda D$, where $R$ is total bitrate, $D$ is total distortion, and $\lambda$ is the Lagrangian multiplier controlling R-D trade-off. \citet{li2022rate} are the pioneer of bit allocation for NVC, who improve the empirical R-D (Rate-Distortion) model from traditional video codec \citep{li2014lambda,li2016lambda} and solve the per-frame Lagrangian multiplier $\lambda$. Other concurrent works adopt simple heuristics for coarse bit allocation \citep{cetin2022flexible,hu2022coarse}.

Most recently, BAO (Bit Allocation using Optimization) \citep{bao2022} proposes to formulate bit allocation as semi-amortized variational inference (SAVI) \citep{kim2018semi,marino2018iterative} and solves it by gradient-based optimization. Specifically, it directly optimizes the variational posterior parameter to be quantized and encoded by gradient ascent, aiming at maximizing the minus overall R-D cost, which is also the evident lowerbound (ELBO). BAO does not rely on any empirical R-D model and thus outperforms previous work. Further, BAO shows its optimality by proving its equivalence to bit allocation with precise R-D model.  

In this paper, we first show that BAO \citep{bao2022} is in fact, sub-optimal due to its implementation. Specifically, we find that it abuses SAVI \citep{kim2018semi,marino2018iterative} on latent with non-factorized variational posterior, which brings incorrect gradient signal during optimization. To solve this problem, we first extend SAVI to non-factorized latent by back-propagating through gradient ascent \citep{domke2012generic}. Then based on that, we correct the sub-optimal bit allocation in BAO to produce true optimal bit allocation for NVC. Furthermore, we propose a computational feasible approximation to such correct but intractable bit allocation method. And we show that our approximation outperforms the incorrect bit allocation (BAO) in terms of R-D performance and bitrate error, and performs better than all other bit allocation methods.

To summarize, our contributions are as follows:
\begin{itemize}
    \item We demonstrate that a previously claimed optimal bit allocation method is actually sub-optimal. We find that its sub-optimality comes from the improper application of SAVI to non-factorized latent.
    \item We present the correct way to conduct SAVI on non-factorized latent by recursively applying back-propagation through gradient ascent. Based on this, we derive the corrected optimal bit allocation algorithm for NVC.
    \item Furthermore, we propose a computational efficient approximation of the optimal bit allocation to make it feasible. Our proposed approach improves the R-D performance and bitrate error over the incorrect bit allocation (BAO), and outperforms all other bit allocation methods for NVC.
\end{itemize}

\section{Preliminaries}
\subsection{Neural Video Compression}

The input of NVC is a GoP (Group of Picture) $\bm{x}_{1:T}$, where $\bm{x}_i\in R^{H\times W}$ is the $i^{th}$ frame with $H\times W$ pixels, and $T$ is the number of frame inside the GoP. Most of the works in NVC follow a latent variable model with temporal autoregressive relationship \citep{yang2020hierarchical}. Specifically, to encode $\bm{x}_i$, we first extract the motion latent $\bm{w}_i=f^{w}_{\phi}(\bm{x}_i,\bm{x}'_i)$ from current frame $\bm{x}_i$ and previous reconstructed frame $\bm{x}'_{i-1}$, where $f^{w}_{\phi}(\cdot)$ is the motion encoder parameterized by $\phi$\footnote{Following previous works in deep generative modeling \citep{kingma2013auto,kim2018semi}, we denote all parameters related to encoder as $\phi$, and all parameters related to decoder and prior as $\theta$.}. Then, we encode the quantized latent $\bmt{w}_i=\lfloor\bm{w}_i\rceil$ with the probability mass function (pmf) estimator $P_{\theta}(\bmt{w}_i|\bmt{w}_{<i},\bmt{y}_{<i})$ parameterized by $\theta$, where $\lfloor\cdot\rceil$ is the rounding. Then, we obtain the residual latent $\bm{y}_i=f_{\phi}^y(\bm{x},\bm{x}',\bmt{w})$, where $f_{\phi}^y(\cdot)$ is the residual encoder. Then, similar to how we treat $\bm{w}_i$, we encode the quantized latent $\bmt{y}_i=\lfloor\bm{y}_i\rceil$ with pmf $P_{\theta}(\bmt{y}_i|\bmt{w}_{\le i},\bmt{y}_{<i})$. Finally, we obtain the reconstructed frame $\bm{x}'_i=g^{x}_{\theta}(\bm{x}'_{i-1},\bmt{w}_i,\bmt{y}_i)$, where $g^{x}_{\theta}(\cdot)$ is the decoder parameterized by $\theta$.

As only the motion latent $\bmt{w}_i$ and residual latent $\bmt{y}_i$ exist in the bitstream, the above process can be simplified as Eq.~\ref{eq:enc} and Eq.~\ref{eq:dec}, where $f_{\phi}(\cdot)$ is the generalized encoder and $g_{\theta}(\cdot)$ is the generalized decoder. The target of NVC is to minimize the per-frame R-D cost $R_i+\lambda_i D_i$ (Eq.~\ref{eq:rd}), where $R_i$ is the bitrate, $D_i$ is the distortion and $\lambda_i$ is the Lagrangian multiplier controlling R-D trade-off. The bitrate $R_i$ and distortion $D_i$ is computed as Eq.~\ref{eq:dec}, where $d(\cdot,\cdot)$ is the distortion metric. And $\lambda_i D_i$ can be further interpreted as the data likelihood term $-\log p_{\theta}(\bm{x}_i|\bmt{w}_{\le i},\bmt{y}_{\le i})$ so long as we treat $\lambda_i D_i$ as the energy function of a Gibbs distribution \citep{minnen2018joint}. Specifically, when $d(\cdot,\cdot)$ is MSE, we can interpret $\lambda_iD_i=-\log p_{\theta}(\bm{x}_i|\bmt{w}_{\le i},\bmt{y}_{\le i})+const$, where $p_{\theta}(\bm{x}_i|\bmt{w}_{\le i},\bmt{y}_{\le i})$ is a Gaussian distribution $\mathcal{N}(\bm{\hat{x}}_i,1/2\lambda_i I)$. 
\begin{align}
    \bm{w}_i = f_{\phi}(\bm{x}_i, \bmt{w}_{<i},\bmt{y}_{<i}),\bm{y}_i = f_{\phi}(\bm{x}_i, \bmt{w}_{\le i}, \bmt{y}_{<i})&\textrm{, where } \bmt{w}_i=\lfloor\bm{w}_i\rceil\textrm{, }\bmt{y}_i=\lfloor\bm{y}_i\rceil \label{eq:enc}\\
    R_i=\log P_{\theta}(\bmt{w}_i,\bmt{y}_i|\bmt{w}_{<i},\bmt{y}_{<i})\textrm{, }D_i=d&(\bm{x}_i,g_{\theta}(\bmt{w}_{\le i},\bmt{y}_{\le i})) \label{eq:dec}\\
    \max -(R_i + \lambda_i D_i&) \label{eq:rd}
\end{align}
On the other hand, NVC is also closely related to Variational Autoencoder (VAE) \citep{kingma2013auto}. As the rounding $\lfloor\cdot\rceil$ is not differentiable, \citet{balle2016end,Theis17} propose to relax it by additive uniform noise (AUN), and replace $\bmt{w}_i=\lfloor\bm{w}_i\rceil$, $\bmt{y}_i=\lfloor\bm{y}_i\rceil$ with $\bmt{w}_i=\bm{w}_i+\mathcal{U}(-0.5,0.5)$, $\bmt{y}_i=\bm{y}_i+\mathcal{U}(-0.5,0.5)$. Under such formulation, the above encoding-decoding process becomes a VAE on graphic model $\bmt{w}_{\le i}, \bmt{y}_{\le i} \rightarrow \bm{x}_i$ with variational posterior as Eq.~\ref{eq:q}, where $\bm{w}_i,\bm{y}_i$ plays the role of variational posterior parameter. Then, minimizing the overall R-D cost (Eq.~\ref{eq:rd}) is equivalent to maximizing the evident lowerbound (ELBO) (Eq.~\ref{eq:elbo}).
\begin{align}
    \hspace{-0.5em}q_{\phi}(\bmt{w}_i|\bm{x}_i, \bmt{w}_{<i}, \bmt{y}_{<i}) = \mathcal{U}(\bm{w}_i-0.5,\bm{w}_i+0.5), q_{\phi}(\bmt{y}_i&|\bm{x}_i, \bmt{w}_{\le i}, \bmt{y}_{<i}) = \mathcal{U}(\bm{y}_i-0.5,\bm{y}_i+0.5) \label{eq:q}\\
    -(R_i + \lambda_i D_i) = \mathbb{E}_{q_{\phi}}[\underbrace{\log P_{\theta}(\bmt{w}_i,\bmt{y}_i|\bmt{w}_{<i},\bmt{y}_{<i})}_{-R_i}&+\underbrace{\log p_{\theta}(\bm{x}_i|\bmt{w}_{\le i},\bmt{y}_{\le i})}_{-\lambda_i D_i}\underbrace{\cancel{-\log q_{\phi}}}_{\textrm{bits-back bitrate: 0}}] \label{eq:elbo}
\end{align}
\subsection{Bit Allocation for Neural Video Compression}
\label{sec:bgba}
It is well known to video coding community that using the same R-D trade-off parameter $\lambda_i$ to optimize R-D cost in Eq.~\ref{eq:rd} for all $T$ frames inside a GoP is suboptimal \citep{li2014lambda,li2016lambda}. This sub-optimality comes from the frame reference structure and is explained in detail by \citet{li2022rate, bao2022}. The target of bit allocation is to maximize the minus of overall R-D cost (ELBO) $\mathcal{L}$ as Eq.~\ref{eq:l0} given the overall R-D trade-off parameter $\lambda_0$, instead of maximizing $\mathcal{L}_i$ of each frame $i$ separately.

The pioneer work of bit allocation in NVC \citep{li2022rate} follows bit allocation for traditional video codec \citep{li2016lambda}. Specifically, it adopts empirical models to approximate the relationship of the rate dependency $\partial R_{i+1}/\partial R_{i}$ and distortion dependency $\partial D_{i+1}/\partial D_{i}$ between frames. Then it takes those models into Eq.~\ref{eq:l0} to solve $\lambda_{1:T}^*$ explicitly as Eq.~\ref{eq:rd1}.\textit{left}. However, its performance heavily relies on the accuracy of empirical models.
\begin{align}
    &\max \mathcal{L} = \sum_{i=1}^{T} \mathcal{L}_{i}\textrm{, where } \mathcal{L}_i = -(R_i + \lambda_0 D_i)\label{eq:l0}\\
    \lambda^{*}_{1:T} \leftarrow \arg \max_{\lambda_{1:T}}& \mathcal{L} (\lambda_{1:T})\textrm{, versus } \bm{w}^{*}_{1:T},\bm{y}^{*}_{1:T} \leftarrow \arg \max_{\bm{w}_{1:T},\bm{y}_{1:T}} \mathcal{L}(\bm{w}_{1:T},\bm{y}_{1:T}) \label{eq:rd1}
\end{align}
On the other hand, BAO \citep{bao2022} does not solve $\lambda_{1:T}^{*}$ explicitly. Instead, it adopts SAVI \citep{kim2018semi,marino2018iterative} to achieve implicit bit allocation. To be specific, it initializes the variational posterior parameter $\bm{w}_{1:T}^0, \bm{y}_{1:T}^0$ from fully amortized variational inference (FAVI) as Eq.~\ref{eq:enc}. Then, it optimizes $\bm{w}_{1:T},\bm{y}_{1:T}$ via gradient ascent to maximize $\mathcal{L}$ as Eq.~\ref{eq:rd1}.\textit{right}. During this procedure, no empirical model is required. BAO further proofs that optimizing Eq.~\ref{eq:rd1}.\textit{right} is equivalent to optimizing Eq.~\ref{eq:rd1}.\textit{left} with precise rate and distortion dependency model $\partial R_{i+1}/\partial R_i,\partial D_{i+1}/\partial D_i$ (See Thm.~1, Thm.~2 in \citet{bao2022}). Thus, BAO claims that it is optimal assuming gradient ascent achieves global maximum. However, in next section, we show that BAO \citep{bao2022} is in fact suboptimal due to its implementation.
\section{Why BAO is Sup-optimal}
\label{sec:baosub}
\begin{figure}[thb]
\centering
 \includegraphics[width=\linewidth]{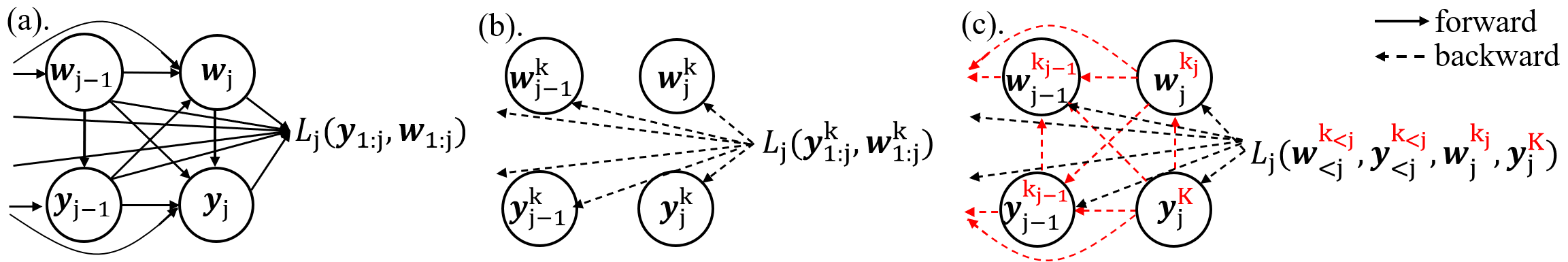}
 \caption{(a). The gradient structure of NVC without SAVI. (b). After $k$ step of SAVI/gradient ascent, the gradient structure of NVC is broken. (c). The proposed approach using back-propagating through gradient ascent. We mark the difference between (b) and (c) in red.}
 \label{fig:grad}
\end{figure}

BAO \citep{bao2022} achieves the SAVI \citep{kim2018semi,marino2018iterative} target in Eq.~\ref{eq:rd1}.\textit{right} by gradient-based optimization. More specifically, its update rule is described as Eq.~\ref{eq:fa_gradw} and Eq.~\ref{eq:fa_grady}, where $K$ is the total number of gradient ascent steps, and $\bm{w}_i^k,\bm{y}_i^k$ is the posterior parameter $\bm{w}_i,\bm{y}_i$ after $k$ steps of gradient ascent. In the original paper of BAO, the authors also find that directly optimizing $\bm{w}_i,\bm{y}_i$ simultaneously by Eq.~\ref{eq:fa_gradw} and Eq.~\ref{eq:fa_grady} performs worse than optimizing $\bm{y}_i$ alone using Eq.~\ref{eq:fa_grady}, but they have not offered any explanation. It is obvious that optimizing $\bm{y}_i$ alone is sub-optimal. However, it is not obvious why jointly optimizing $\bm{w}_i, \bm{y}_i$ with Eq.~\ref{eq:fa_gradw} and Eq.~\ref{eq:fa_grady} fails.
\begin{align}
     \bm{w}^{k+1}_i \leftarrow \bm{w}^k_i + \alpha\frac{ d\mathcal{L}(\bm{w}^k_{1:T},\bm{y}^k_{1:T})}{d \bm{w}^k_i}\textrm{, where }\frac{d \mathcal{L}(\bm{w}^k_{1:T},\bm{y}^k_{1:T})}{d \bm{w}^k_i}=\sum_{j=i}^{T}\frac{\partial \mathcal{L}_{j}(\bm{w}^k_{1:j},\bm{y}^k_{1:j})}{\partial \bm{w}^k_i}\label{eq:fa_gradw}\\
     \bm{y}^{k+1}_i \leftarrow \bm{y}^k_i + \alpha\frac{d \mathcal{L}(\bm{w}^k_{1:T},\bm{y}^k_{1:T})}{d \bm{y}^k_i}\textrm{, where }\frac{d \mathcal{L}(\bm{w}^k_{1:T},\bm{y}^k_{1:T})}{d \bm{y}^k_i}=\sum_{j=i}^{T}\frac{\partial \mathcal{L}_{j}(\bm{w}^k_{1:j},\bm{y}^k_{1:j})}{\partial\bm{y}^k_i}
     \label{eq:fa_grady}
\end{align}
In fact, the update rule in Eq.~\ref{eq:fa_gradw} and Eq.~\ref{eq:fa_grady} is exactly the SAVI \citep{kim2018semi,marino2018iterative} when $\bm{w}_i,\bm{y}_i$ fully factorizes (e.g. the full factorization used in mean-field \citep{blei2017variational}). However, in NVC the $\bm{w}_i,\bm{y}_i$ has complicated auto-regressive relationships (See Eq.~\ref{eq:enc} and Fig.~\ref{fig:grad}.(a)). Abusing SAVI on non-factorized latent causes gradient error in two aspects: (1). The total derivative $d \mathcal{L}/d\bm{w}_i,d \mathcal{L}/d\bm{y}_i$ is incomplete. (2). The total derivative $d \mathcal{L}/d\bm{w}_i,d \mathcal{L}/d\bm{y}_i$ and partial derivative $\partial \mathcal{L}_j/\partial\bm{w}_i,\partial\mathcal{L}_j/\partial\bm{y}_i$ is evaluated at wrong value. In next two sections, we elaborate those two issues with $\bm{w}_i$ related equations in main text and $\bm{y}_i$ related equations in Appendix.~\ref{app:cf}.
\subsection{Incomplete Total derivative evaluation}
\label{sec:itde}
According to the latent generation procedure described by Eq.~\ref{eq:enc} and Eq.~\ref{eq:dec}, we draw the computational graph to describe the latent dependency as Fig.~\ref{fig:grad}.(a). Based on that, we expand the total derivative $d\mathcal{L}/d\bm{w}_i,d\mathcal{L}/d\bm{y}_i$ as Eq.~\ref{eq:incwg} and Eq.~\ref{eq:incyg}.
\begin{align}
\frac{d \mathcal{L}(\bm{w}_{1:T},\bm{y}_{1:T})}{d \bm{w}_i}=&\sum_{j=i}^{T}\frac{d \mathcal{L}_j(\bm{w}_{1:j},\bm{y}_{1:j})}{d \bm{w}_i}\notag\\
\frac{d \mathcal{L}_j(\bm{w}_{1:j},\bm{y}_{1:j})}{d \bm{w}_i}=&\underbrace{\sum_{l=i+1}^{j}\frac{\partial \bm{w}_l}{\partial\bm{w}_i}\frac{d \mathcal{L}_{j}(\bm{w}_{1:j},\bm{y}_{1:j})}{d \bm{w}_l}+\sum_{l=i}^{j}\frac{\partial \bm{y}_l}{\partial\bm{w}_i}\frac{d \mathcal{L}_{j}(\bm{w}_{1:j},\bm{y}_{1:j})}{d \bm{y}_l}}_{\textrm{ignored by BAO}}+\underbrace{\frac{\partial \mathcal{L}_j(\bm{w}_{1:j},\bm{y}_{1:j})}{\partial \bm{w}_i}}_{\textrm{considered by BAO}}\label{eq:incwg}
\end{align}
As shown in Eq.~\ref{eq:fa_gradw}, Eq.~\ref{eq:fa_grady} and Fig.~\ref{fig:grad}.(b), BAO \citep{bao2022} treats the total derivative $d\mathcal{L}/d\bm{w}_i,d\mathcal{L}/d\bm{y}_i$ as the sum of the frame level partial derivative $\partial \mathcal{L}_j/\partial\bm{w}_i,\partial \mathcal{L}_j/\partial\bm{y}_i$, which is the direct contribution of frame $i^{th}$ latent $\bm{w}_i,\bm{y}_i$ to $j^{th}$ frame's R-D cost $\mathcal{L}_j$ (as marked in Eq.~\ref{eq:incwg} and Eq.~\ref{eq:incyg}). This incomplete evaluation of gradient signal brings sub-optimality. Further, it is not possible to correct BAO by simply including other parts of gradient into consideration. As BAO jointly updates all the latent $\bm{w}_{1:T},\bm{y}_{1:T}$, the relationship of Eq.~\ref{eq:dec} only holds for the initial latent parameters $\bm{w}_{1:T}^0,\bm{y}_{1:T}^0$ produced by FAVI. And this important relationship is broken for parameters $\bm{w}_{1:T}^k,\bm{y}_{1:T}^k$ after $k\ge1$ steps of update.

\subsection{Incorrect Value to Evaluate Gradient}
\label{sec:ipde}
As shown in Eq.~\ref{eq:fa_gradw} and Eq.~\ref{eq:fa_grady}, BAO \citep{bao2022} simultaneously updates all the posterior parameter $\bm{w}_{1:T},\bm{y}_{1:T}$ with gradient evaluated at the same gradient ascent step $\bm{w}_{1:T}^k,\bm{y}_{1:T}^k$. However, as we show later in Sec.~\ref{sec:savi2} and Fig.~\ref{fig:grad}.(c), this is sub-optimal as all the descendant latent $\bm{w}_{>i},\bm{y}_{\ge i}$ of $\bm{w}_i$ should already complete all $K$ steps of gradient ascent before the gradient of $\bm{w}_i$ is evaluated. Moreover, $\bm{w}_{>i},\bm{y}_{\ge i}$ should be initialized by FAVI using precedents latent. Similar rule applies to $\bm{y}_i$. Specifically, the correct value to evaluate the gradient is as Eq.~\ref{eq:incygw} and Eq.~\ref{eq:incygy}, where $\bm{w}_i^{k_i}$ denotes the latent $\bm{w}_i$ after $k_i$ steps of update, and $\bm{y}_i^{k'_j}$ denotes the latent $\bm{y}_i$ after $k'_i$ steps of update.
\begin{align}
     \bm{w}^{k_i+1}_i \leftarrow \bm{w}^{k_i}_i + \alpha\frac{ d\mathcal{L}(\bm{w}_1^{k_1},...,\bm{w}_i^{k_i},\bm{w}^K_{>i},\bm{y}_1^{k'_1},...,\bm{y}_{i-1}^{k'_{i-1}},\bm{y}^K_{\ge i})}{d \bm{w}^{k_i}_i},\notag\\\textrm{where }\bm{w}_{>i}^0,\bm{y}_{\ge i}^0=f(\bm{x},\bm{w}_1^{k_1},...,\bm{w}_i^{k_i},\bm{y}_1^{k'_1},...,\bm{y}_{i-1}^{k'_{i-1}})\label{eq:incygw}
\end{align}
Similar to the incomplete total derivative evaluation, this problem does not have a simple solution. In next section, we show how to correct both of the above-mentioned issues by recursively applying back-propagating through gradient ascent \citep{domke2012generic}.

\section{Correcting the Sub-optimal Bit Allocation}
\label{sec:cba}
In this section, we first extend the generic SAVI \citet{kim2018semi,marino2018iterative} to 2-level non-factorized latent. Then we further extend this result to latent with any dependency that can be described by a DAG (Directed Acyclic Graph). And finally, we correct the sub-optimal bit allocation by applying the result in DAG latent to NVC.
\subsection{SAVI on 2-level non-factorized latent}
\label{sec:savi2}
In this section, we extend the SAVI on 1-level latent \citep{kim2018semi} to 2-level non-factorized latent. We denote $\bm{x}$ as evidence, $\bm{w}$ as the variational posterior parameter of the first level latent $\bmt{w}$, $\bm{y}$ as the variational posterior parameter of the second level latent $\bmt{y}$, and the ELBO to maximize as $\mathcal{L}(\bm{w},\bm{y})$. The posterior $q(\bmt{w},\bmt{y}|\bm{x})$ factorizes as $q(\bmt{w}|\bm{x})q(\bmt{y}|\bmt{w},\bm{x})$, which means that $\bm{y}$ depends on $\bm{w}$. Given $\bm{w}$ is fixed, we can directly follow \citet{kim2018semi,marino2018iterative} to optimize $\bm{y}$ to maximize ELBO by SAVI. However, it requires some tricks to optimize $\bm{w}$.

\begin{minipage}[t]{.42\textwidth} %
\vspace{0pt}
\IncMargin{1.0em}
\begin{algorithm}[H]
\DontPrintSemicolon
\caption{SAVI on 2-level Latent}\label{alg:solve-2}
\textbf{procedure} solve-2-level($\bm{x},\bm{w}^{k}$)\;
$\quad$initialize $\bm{w}^0\leftarrow f(\bm{x})$ from FAVI\;
$\quad$\textbf{for} $k=0,...,K-1$ \textbf{do}\;
$\quad\quad \frac{d\mathcal{L}(\bm{w}^k,\bm{y}^K)}{d\bm{w}^k}=\textrm{grad-2-level}(\bm{x},\bm{w}^k)$\;
$\quad\quad \bm{w}^{k+1}\leftarrow\bm{w}^k+\alpha\frac{d\mathcal{L}(\bm{w}^k,\bm{y}^K)}{d\bm{w}^k}$\;
$\quad$\textbf{return} $\bm{w}^{K},\bm{y}^K$\;
\BlankLine
\textbf{procedure} grad-2-level($\bm{x},\bm{w}^{k}$)\;
$\quad\bm{y}^0\leftarrow f(\bm{x},\bm{w}^{k})$ from FAVI\;
$\quad$\textbf{for} $k'=0,...,K-1$ \textbf{do}\;
$\quad\quad \bm{y}^{k'+1}\leftarrow\bm{y}^{k'}+\alpha\frac{d\mathcal{L}(\bm{w}^{k},\bm{y}^{k'})}{d\bm{y}^{k'}}$\;
$\quad\overleftarrow{\bm{w}}\leftarrow\frac{\partial \mathcal{L}(\bm{w}^{k},\bm{y}^K)}{\partial \bm{w}^{k}}$\;
$\quad\overleftarrow{\bm{y}^K}\leftarrow\frac{d\mathcal{L}(\bm{w}^{k},\bm{y}^K)}{d\bm{y}^K}$\;
$\quad$\textbf{for} $k'=K-1,...,0$ \textbf{do}\;
$\quad\quad\overleftarrow{\bm{w}}\leftarrow\overleftarrow{\bm{w}}+\alpha\frac{\partial^2 \mathcal{L}(\bm{w}^{k},\bm{y}^{k'})}{\partial \bm{w}^{k}\partial\bm{y}^{k'}}\overleftarrow{\bm{y}^{k'+1}}$\;
$\quad\quad\overleftarrow{\bm{y}^{k'}}\leftarrow\overleftarrow{\bm{y}^{k'}}+\alpha\frac{\partial^2 \mathcal{L}(\bm{w}^{k},\bm{y}^{k'})}{\partial \bm{y}^{k'}\partial\bm{y}^{k'}}\overleftarrow{\bm{y}^{k'+1}}$\;
$\quad\overleftarrow{\bm{w}}=\overleftarrow{\bm{w}}+\frac{\partial \bm{y}^0}{\partial\bm{w}^k}\overleftarrow{\bm{y}^{0}}$\;
$\quad$\textbf{return} $\frac{d\mathcal{L}(\bm{w}^k,\bm{y}^K)}{d\bm{w}^k}=\overleftarrow{\bm{w}}$\; 
\end{algorithm}
\end{minipage}
\begin{minipage}[t]{.58\textwidth} %
\vspace{0pt}
\IncMargin{1.0em}
\begin{algorithm}[H]
\DontPrintSemicolon
\caption{SAVI on DAG Latent}\label{alg:solve-dag}
\textbf{procedure} solve-dag($\bm{x}$)\;
$\quad$sort $\bm{y}_1,...,\bm{y}_N$ in topological order\;
$\quad$\textbf{for} $\bm{y}_j$ with parent $\mathcal{P}(\bm{y}_j)=\varnothing$\;
$\quad\quad$ add $\bm{y}_j$ to fake node $\bm{y}_0$'s children $\mathcal{C}(\bm{y}_0)$\;
$\quad$grad-dag($\bm{x},\bm{y}_0^0$)\;
$\quad$\textbf{return} $\bm{y}_1^K,...,\bm{y}_N^K$\;
\BlankLine
\textbf{procedure} grad-dag($\bm{x},\bm{y}_0^{k_0},...,\bm{y}_i^{k_i}$)\;
$\quad$\textbf{for} $\bm{y}_{j}\in\mathcal{C}(\bm{y}_i)$ in topological order \textbf{do}\;
$\quad\quad$ $\bm{y}_{j}^0\leftarrow f(\bm{x},\bm{y}_0^{k_0},...,\bm{y}_{<j}^{k_{<j}})$ from FAVI\;
$\quad\quad$ \textbf{for} $k_j=0,...,K-1$ \textbf{do}\;
$\quad\quad\quad \frac{d\mathcal{L}(\bm{y}_0^{k_0},...,\bm{y}_j^{k_j},\bm{y}_{>j}^K)}{d\bm{y}_{j}^{k_j}} \leftarrow \textrm{grad-dag}(\bm{x},\bm{y}_0^{k_0},...,\bm{y}_j^{k_j})$\;
$\quad\quad\quad\bm{y}_{j}^{k_j+1}\leftarrow\bm{y}_{j}^{k_j}+ \alpha\frac{d\mathcal{L}(\bm{y}_0^{k_0},...,\bm{y}_j^{k_j},\bm{y}_{>j}^K)}{d\bm{y}_{j}^{k_j}}$\;
$\quad\overleftarrow{\bm{y}_i}\leftarrow \frac{\partial\mathcal{L}(\bm{y}_0^{k_0},...,\bm{y}_i^{k_i},\bm{y}_{>i}^K)}{\partial \bm{y}_i^{k_i}}$\;
$\quad$\textbf{for} $\bm{y}_{j}\in\mathcal{C}(\bm{y}_i)$ \textbf{do}\;
$\quad\quad\overleftarrow{\bm{y}_j}\leftarrow\bm{0},\overleftarrow{\bm{y}_{j}^K}\leftarrow\frac{d \mathcal{L}(\bm{y}_0^{k_0},...,\bm{y}_i^{k_i},\bm{y}_{>i}^K)}{d\bm{y}_{j}^K}$\;
$\quad\quad$\textbf{for} $k_j=K-1,...,0$  \textbf{do}\;
$\quad\quad\quad\overleftarrow{\bm{y}_j}\leftarrow\overleftarrow{\bm{y}_j}+\alpha \frac{\partial^2 \mathcal{L}(\bm{y}_0^{k_0},...,\bm{y}_j^{k_j},\bm{y}_{>j}^K)}{\partial\bm{y}_i^{k_i}\partial\bm{y}_{j}^{k_j}}\overleftarrow{\bm{y}_{j}^{k_j+1}}$\;
$\quad\quad\quad\overleftarrow{\bm{y}_j^{k_j}}\leftarrow\overleftarrow{\bm{y}_j^{k_j+1}}+\alpha \frac{\partial^2 \mathcal{L}(\bm{y}_0^{k_0},...,\bm{y}_j^{k_j},\bm{y}_{>j}^K)}{\partial\bm{y}_j^{k_j}\partial\bm{y}_{j}^{k_j}}\overleftarrow{\bm{y}_{j}^{k_j+1}}$\;
$\quad\quad\overleftarrow{\bm{y}_i}\leftarrow \overleftarrow{\bm{y}_i} + \overleftarrow{\bm{y}_j}+\frac{\partial \bm{y}_j^0}{\partial \bm{y_i^{k_i}}}\overleftarrow{\bm{y}_j^{0}}$\;
$\quad$\textbf{return}$\frac{d\mathcal{L}(\bm{y}_0^{k_0},...,\bm{y}_i^{k_i},\bm{y}_{>i}^K)}{d \bm{y}_i^{k_i}}=\overleftarrow{\bm{y}_i}$\;
\end{algorithm}
\end{minipage}

The intuition is, we do not want to find a $\bm{w}$ that maximizes $\mathcal{L}(\bm{w},\bm{y})$ given a fixed $\bm{y}$ (or we have the gradient issue described in Sec.~\ref{sec:baosub}). Instead, we want to find a $\bm{w}$, whose $\max_{\bm{y}}\mathcal{L}(\bm{w},\bm{y})$ is maximum. This translates to the optimization problem as Eq.~\ref{eq:opt2}. In fact, Eq.~\ref{eq:opt2} is a variant of setup in back-propagating through gradient ascent \citep{samuel2009learning,domke2012generic}. The difference is, our $\bm{w}$ also contributes directly to optimization target $\mathcal{L}(\bm{w},\bm{y})$. From this perspective, Eq.~\ref{eq:opt2} is more closely connected to \citet{kim2018semi}, if we treat $\bm{w}$ as the model parameter and $\bm{y}$ as latent. 
\begin{align}
    \bm{w}\leftarrow \arg \max_{\bm{w}} \mathcal{L}(\bm{w},\bm{y}^*(\bm{w}))\textrm{, where } \bm{y}^*(\bm{w})\leftarrow \arg\max_{\bm{y}} \mathcal{L}(\bm{w},\bm{y})
    \label{eq:opt2}
\end{align}
And as SAVI on 1-level latent \citep{kim2018semi,marino2018iterative}, we need to solve Eq.~\ref{eq:opt2} using gradient ascent. Specifically, denote $\alpha$ as step size (learning rate), $K$ as the total gradient ascent steps, $\bm{w}^k$ as the $\bm{w}$ after $k$ step update,  $\bm{y}^{k'}$ as the $\bm{y}$ after $k'$ step update, and $f(.)$ as FAVI procedure generating initial posterior parameters $\bm{w}^0,\bm{y}^0$, the optimization problem as Eq.~\ref{eq:opt2} translates into the update rule as Eq.~\ref{eq:opt2grad}. Eq.~\ref{eq:opt2grad} is the guidance for designing optimization algorithm, and it also explains why the gradient of BAO \citep{bao2022} is evaluated at wrong value (See Sec.~\ref{sec:ipde}).
\begin{align}
    \bm{w}^{k+1}\leftarrow \bm{w}^{k}+\alpha\frac{d\mathcal{L}(\bm{w}^k,\bm{y}^{K})}{d\bm{w}^k},
    \bm{y}^{k'+1}\leftarrow \bm{y}^{k'}+\alpha\frac{d\mathcal{L}(\bm{w}^k,\bm{y}^{k'})}{d\bm{y}^{k'}}\textrm{, where } \bm{y}^0 = f(\bm{x},\bm{w}^k)\label{eq:opt2grad}
\end{align}
To solve Eq.~\ref{eq:opt2grad}, we note that although $d\mathcal{L}(\bm{w}^k,\bm{y}^{k'})/d\bm{y}^{k'}$ is directly computed, $d\mathcal{L}(\bm{w}^k,\bm{y}^{K})/d\bm{w}^{k}$ is not straightforward. Resorting to previous works \citep{samuel2009learning,domke2012generic} in implicit differentiation and extending the results in \citet{kim2018semi} from model parameters to variational posterior parameters, we implement Eq.~\ref{eq:opt2grad} as Alg.~\ref{alg:solve-2}. Specifically, we first initialize $\bm{w}^0$ from FAVI. Then we conduct gradient ascent on $\bm{w}$ with gradient $d\mathcal{L}(\bm{w}^k,\bm{y}^K)/d\bm{w}^{k}$ computed from the procedure grad-2-level($\bm{x},\bm{w}^k$).   And inside grad-2-level($\bm{x},\bm{w}^k$), $\bm{y}$ is also updated by gradient ascent, the above procedure corresponds to Eq.~\ref{eq:opt2grad}. The key of Alg.~\ref{alg:solve-2} is the evaluation of gradient $d\mathcal{L}(\bm{w}^k,\bm{y}^K)/d\bm{w}^{k}$. Formally, we have:
\begin{theorem}
\label{th:2l}
After \textup{grad-2-level($\bm{x},\bm{w}^k$)} of Alg.~\ref{alg:solve-2} executes, we have the return value $d \mathcal{L}(\bm{w}^k,\bm{y}^K)/d\bm{w}^k=\overleftarrow{\bm{w}}$. (See proof in Appendix.~\ref{app:pf}.)
\end{theorem}
\subsection{SAVI on DAG-defined Non-factorized Latent}
\label{sec:savidag}
In this section, we extend the result from previous section to SAVI on general non-factorized latent with dependency described by any DAG. This DAG is the computational graph during network inference, and it is also the directed graphical model (DGM) \citep{koller2009probabilistic} defining the factorization of latent variables during inference. This is the general case covering all dependency that can be described by DGM. This extension is necessary to perform SAVI on latent with complicated dependency (e.g. bit allocation of NVC).

Similar to the 2-level latent setup, we consider performing SAVI on $N$ variational posterior parameter $\bm{y}_1,...,\bm{y}_N$ with their dependency defined by a computational graph $\mathcal{G}$, i.e., their corresponding latent variable $\bmt{y}_1,...,\bmt{y}_N$'s posterior distribution factorizes as $\mathcal{G}$. Specifically, we denote $\bm{y}_j\in\mathcal{C}(\bm{y}_i),\bm{y}_i\in\mathcal{P}(\bm{y}_j)$ if an edge exists from $\bm{y}_i$ to $\bm{y}_j$. This indicates that $\bmt{y}_j$ conditions on $\bmt{y}_i$. Without loss of generality, we assume $\bm{y}_1,...,\bm{y}_N$ is sorted in topological order. This means that if $\bm{y}_j\in\mathcal{C}(\bm{y}_i),\bm{y}_i\in\mathcal{P}(\bm{y}_j)$, then $i<j$. Each latent is optimized by $K$-step gradient ascent, and $\bm{y}_i^{k_i}$ denotes the latent $\bm{y}_i$ after $k_i$ steps of update. Then, similar to 2-level latent, we have the update rule as Eq.~\ref{eq:optdaggrad}:
\begin{align}
\bm{y}^{k_i+1}_i\leftarrow \bm{y}^{k_i}_i+\alpha\frac{d\mathcal{L}(\bm{y}_1^{k_1},...,\bm{y}_i^{k_i},\bm{y}_{>i}^K)}{d\bm{y}^{k_i}}\textrm{, where }\bm{y}_{>i}^0 = f(\bm{x},\bm{y}_1^{k_1},...,\bm{y}_i^{k_i})
\label{eq:optdaggrad}
\end{align}
, which can be translated into Alg.~\ref{alg:solve-dag}. Specifically, we first sort the latent in topological order. Then, we add a fake latent $\bm{y}_0$ to the front of all $\bm{y}$s. Its children are all the $\bm{y}s$ with 0 in-degree. Then, we can solve the SAVI on $\bm{y}_1,...,\bm{y}_N$ using gradient ascent by executing the procedure grad-dag($\bm{x},\bm{y}_0^{k_0},...,\bm{y}_i^{k_i}$) in Alg.~\ref{alg:solve-dag} recursively. Inside procedure grad-dag($\bm{x},\bm{y}_0^{k_0},...,\bm{y}_i^{k_i}$), the gradient to update $\bm{y}_i$ relies on the convergence of its children $\bm{y}_{j}\in\mathcal{C}(\bm{y}_i)$, which is implemented by the recursive depth-first search (DFS) in line 11. And upon the completion of procedure grad-dag($\bm{x},\bm{y}_0^0$), all the latent converges to $\bm{y}_1^K,...,\bm{y}_N^K$. Similar to the 2-level latent case, the key of Alg.~\ref{alg:solve-dag} is the evaluation of gradient $d\mathcal{L}(\bm{y}_0^{k_0},...,\bm{y}_i^{k_i},\bm{y}_{>i}^K)/d \bm{y}_i^{k_i}$. Formally, we have:
\begin{theorem}
\label{th:dag}
After the procedure \textup{grad-dag($\bm{x},\bm{y}_0^{k_0},...,\bm{y}_i^{k_i}$)} in Alg.~\ref{alg:solve-dag} executes, we have the return value $d\mathcal{L}(\bm{y}_0^{k_0},...,\bm{y}_i^{k_i},\bm{y}_{>i}^K)/d \bm{y}_i^{k_i}=\overleftarrow{\bm{y}_i}$. (See proof in Appendix.~\ref{app:pf}.)
\end{theorem}
To better understand how Alg.~\ref{alg:solve-dag} works, we provide a detailed example in Fig.~\ref{fig:eg} of Appendix.~\ref{app:eg}.
\subsection{Correcting the Sub-optimal Bit Allocation using SAVI on DAG}
\label{sec:savinvc}
With the result in previous section, correcting BAO \citep{bao2022} seems to be trivial. We only need to sort the latent in topological order as $\bm{w}_1,\bm{y}_1,...,\bm{w}_T,\bm{y}_T$, and run Alg.~\ref{alg:solve-dag} to obtain the optimized latent parameters $\bm{w}_1^K,\bm{y}_1^K,...,\bm{w}_T^K,\bm{y}_T^K$. And the gradient $d\mathcal{L}(\bm{y}_0^{k_0},...,\bm{y}_i^{k_i},\bm{y}_{>i}^K)/d \bm{y}_i^{k_i}$ computed in Alg.~\ref{alg:solve-dag} resolves the issue of BAO described in Sec.~\ref{sec:itde} and Sec.~\ref{sec:ipde}. However, an evident problem is the temporal complexity. Given the latent number $N$ and gradient ascent step number $K$, Alg.~\ref{alg:solve-dag} has temporal complexity of $\Theta(K^N)$. NVC with GoP size $10$ has approximately $N=20$ latent, and the SAVI on NVC \citep{bao2022} takes around $K=2000$ step to converge. For bit allocation, the complexity of Alg.~\ref{alg:solve-dag} is $\approx 2000^{20}$, which is intractable. On the other hand, BAO's complexity is reasonable ($\Theta(KN)\approx4\times10^{4}$). Thus, in next section, we provide a feasible approximation to such intractable corrected bit allocation.

\subsection{Feasible Approximation to the Corrected Bit Allocation}
\label{sec:approx}
In order to solve problem with practical size such as bit allocation on NVC, we provide an approximation to the SAVI \citep{kim2018semi,marino2018iterative} on DAG described in Sec.~\ref{sec:savidag}. The general idea is that, when being applied to bit allocation of NVC, the accurate SAVI on DAG (Alg.~\ref{alg:solve-dag}) satisfies both requirement on gradient signal described in Sec.~\ref{sec:itde} and Sec.~\ref{sec:ipde}. We can not make it tractable without breaking them. Thus, we break one of them and achieve a reasonable complexity, while maintain a superior performance compared with BAO \citep{bao2022}.

We consider the approximation in Eq.~\ref{eq:approx} which breaks the requirement for gradient evaluation in Sec.~\ref{sec:ipde}. Based on Eq.~\ref{eq:approx} and the requirement in Sec.~\ref{sec:itde}, we design an approximation of accurate SAVI as Alg.~\ref{alg:solve-adag}. When being applied to bit allocation in NVC, it satisfies the gradient requirement in Sec.~\ref{sec:itde} while maintaining a temporal complexity of $\Theta(KN)$ as BAO.
\begin{align}
\frac{d\mathcal{L}(\bm{y}_0^{k_0},...,\bm{y}_i^{k_i},\bm{y}_{>i}^K)}{d \bm{y}_i^{k_i}}\approx\frac{d\mathcal{L}(\bm{y}_0^{k_0},...,\bm{y}_i^{k_i},\bm{y}_{>i}^0)}{d \bm{y}_i^{k_i}} \label{eq:approx}
\end{align}
Specifically, with the approximation in Eq.~\ref{eq:approx}, the recurrent gradient computation in Alg.~\ref{alg:solve-dag} becomes unnecessary as the right hand side of Eq.~\ref{eq:approx} does not require $\bm{y}_{>i}^K$. However, to maintain the dependency of latent described in Sec.~\ref{sec:itde}, as Alg.~\ref{alg:solve-dag}, we still need to ensure that the children node $\bm{y}_{j}\in \mathcal{C}(\bm{y}_i)$ are re-initialized by FAVI every-time when $\bm{y}_i$ is updated. Therefore, a reasonable approach is to traverse the graph in topological order. We keep the children node $\bm{y}_j$ untouched until all its parent node $\bm{y}_i\in\mathcal{P}(\bm{y}_j)$'s gradient ascent is completed and $\bm{y}_i^K$ is known. And the resulting approximate SAVI algorithm is as Alg.~\ref{alg:solve-adag}. When applied to bit allocation, it satisfies the gradient requirement in Sec.~\ref{sec:itde}, and as BAO, its temporal complexity is $\Theta(KN)$. 
\begin{minipage}[t]{.44\textwidth} %
\vspace{0pt}
\IncMargin{1.0em}
\begin{algorithm}[H]
\DontPrintSemicolon
\caption{BAO on DAG Latent}\label{alg:solve-bao}
\textbf{procedure} solve-bao($\bm{x}$)\;
$\quad\bm{y}_1^0,...,\bm{y}_N^0\leftarrow f(\bm{x})$ from FAVI\;
$\quad$\textbf{for} $k =0,...,K-1$ \textbf{do}\;
$\quad\quad$ \textbf{for} $i=1,...,N$ \textbf{do}\;
$\quad\quad\quad \bm{y}_i^{k+1}\leftarrow\bm{y}_i^k+\alpha\frac{\partial\mathcal{L}(\bm{y}_1^k,...,\bm{y}_N^k)}{\partial\bm{y}_i^k}$\;
$\quad$\textbf{return} $y_1^K,...,y_N^K$\;
\end{algorithm}
\end{minipage}
\begin{minipage}[t]{.55\textwidth} %
\vspace{0pt}
\IncMargin{1.0em}
\begin{algorithm}[H]
\DontPrintSemicolon
\caption{Approximate SAVI on DAG latent}\label{alg:solve-adag}
\textbf{procedure} solve-approx-dag($\bm{x}$)\;
$\quad$sort $\bm{y}_1,...,\bm{y}_N$ in topological order\;
$\quad$\textbf{for} $i=1,...,N$ \textbf{do}\;
$\quad\quad \bm{y}_i^0,...,\bm{y}_N^0\leftarrow f(\bm{x},\bm{y}_{<i}^K)$ from FAVI\;
$\quad\quad$ \textbf{for} $k=0,...,K-1$ \textbf{do}\;
$\quad\quad\quad\frac{d\mathcal{L}(\bm{y}_{<i}^K,\bm{y}_i^k,\bm{y}_{>i}^K)}{d\bm{y}_i^k}\approx\frac{d\mathcal{L}(\bm{y}_{<i}^K,\bm{y}_i^k,\bm{y}_{>i}^0)}{d\bm{y}_i^k}$\;
$\quad\quad\quad \bm{y}_i^{k+1}\leftarrow\bm{y}_i^k+\alpha\frac{d\mathcal{L}(\bm{y}_{<i}^K,\bm{y}_i^k,\bm{y}_{>i}^K)}{d\bm{y}_i^k}$\;
$\quad$\textbf{return} $y_1^K,...,y_N^K$\;
\end{algorithm}
\end{minipage} %

To better understand BAO \citep{bao2022} in SAVI context, we rewrite it by general SAVI notation instead of NVC notation in Alg.~\ref{alg:solve-bao}. We highlight the difference between BAO (Alg.~\ref{alg:solve-bao}) \citep{bao2022}, the accurate SAVI on DAG latent (Alg.~\ref{alg:solve-dag}) and the approximate SAVI on DAG latent (Alg.~\ref{alg:solve-adag}) from several aspects:
\begin{itemize}
    \item \textbf{Graph Traversal Order}: BAO performs gradient ascent on $\bm{y}_{1:T}$ all together. The accurate SAVI only updates $\bm{y}_i$ when $\bm{y}_{>i}$'s update is complete and $\bm{y}_{>i}^K$ is known. The approximate SAVI only updates $\bm{y}_i$ when $\bm{y}_{<i}$'s update is complete and $\bm{y}_{<i}^K$ is known. 
    \item \textbf{Gradient Correctness}: When being applied to bit allocation in NVC, BAO violates the gradient rule in Sec.~\ref{sec:itde} and Sec.~\ref{sec:ipde}, accurate SAVI satisfies both rules, approximate SAVI satisfies Sec.~\ref{sec:itde} and violates Sec.~\ref{sec:ipde}.
    \item \textbf{Temporal Complexity}: With the latent number $N$ and steps of gradient ascent $K$, the complexity of BAO is $\Theta(KN)$, the complexity of accurate SAVI is $\Theta(K^N)$ and the complexity of approximate SAVI is $\Theta(KN)$.
\end{itemize}

Then we can simply apply Alg.~\ref{alg:solve-adag} to bit allocation in NVC to obtain a feasible approximation of the corrected optimal bit allocation. And in Sec.~\ref{sec:rd}, we empirically show that our approximation improves the R-D performance over BAO \citep{bao2022} with even smaller number of updates.

\section{Related Work: Bit Allocation \& SAVI for Neural Compression}
\citet{li2022rate} are the pioneer of bit allocation for NVC and their work is elaborated in Sec.~\ref{sec:bgba}. Other recent works that consider bit allocation for NVC only adopt simple heuristic such as inserting $1$ high quality frame per $4$ frames \citep{hu2022coarse,cetin2022flexible}. On the other hand, OEU \citep{lu2020content} is also recognised as frame-level bit allocation while its performance is inferior than BAO \citep{bao2022}. BAO is the most recent work with best R-D performance. It is elaborated in Sec.~\ref{sec:bgba} and Sec.~\ref{sec:baosub}, and corrected in the previous section.

Semi-Amortized Variational Inference (SAVI) is proposed by \citet{kim2018semi,marino2018iterative}. The idea is that works following \citet{kingma2013auto} use fully amortized inference parameter $\phi$ for all data, which leads to the amortization gap \citep{cremer2018inference}. SAVI reduces this gap by optimizing the variational posterior parameter after initializing it with inference network. It adopts back-propagating through gradient ascent \citep{domke2012generic} to evaluate the gradient of model parameters. We adopt a similar method to extend SAVI to non-factorized latent. When applying SAVI to practical neural codec, researchers abandon the nested model parameter update for 
efficiency. Prior works \citep{djelouah2019content,yang2020improving,zhao2021universal,ce2022} adopt SAVI to boost R-D performance and achieve variable bitrate in image compression. And BAO \citep{bao2022} is the first to consider SAVI for bit allocation.

\section{Experiments}
\subsection{Experimental Settings}
We implement our approach in PyTorch 1.9 with CUDA 11.2, and run the experiments on NVIDIA(R) A100 GPU. Most of the other settings are intentionally kept the same as BAO \citep{bao2022}. Specifically, we adopt HEVC Common Testing Condition (CTC) \citep{bossen2013common} and UVG dataset \citep{mercat2020uvg}. And we measure the R-D performance in Bjontegaard-Bitrate (BD-BR) and BD-PSNR \citep{bjontegaard2001calculation}. For baseline NVC \citep{lu2019dvc,li2021deep}, we adopt the official pre-trained models. And we select target $\lambda_0=\{256,512,1024,2048\}$. For gradient ascent, we adopt Adam \citep{kingma2014adam} optimizer with $lr=1\times10^{-3}$. We set the gradient ascent step $K=2000$ for the first frame and $K=400$ for other frames. More details are presented in Appendix.~\ref{app:impl}.

\subsection{Quantitative Results}
\label{sec:rd}
As shown in Tab.~\ref{tab:bdbr}, our method consistently improves the R-D performance in terms of BD-BR over BAO \citep{bao2022} on both baseline methods and all datasets. Moreover, this improvement is especially significant (more than 10\% in BD-BR) when the baseline is DCVC \citep{li2021deep}. And both BAO and our proposed correction significantly outperform other approaches. It is also noteworthy that with our bit allocation, DVC (the SOTA method in 2019) already outperforms DCVC (the SOTA method in 2021) by large margin (See the red solid line and black dash line in Fig.~\ref{fig:bdbrd}).

\begin{minipage}{\textwidth}
\vspace{7pt}
\begin{minipage}[t]{0.7\textwidth}
\vspace{0pt}
\centering
\begin{small}
\begin{tabular}{@{}llllll@{}}
\toprule
 & \multicolumn{5}{c}{BD-BR (\%) $\downarrow$} \\ \cmidrule(rr){2-6}
Method & Class B & Class C & Class D & Class E & UVG  \\ \midrule
\multicolumn{3}{@{}l@{}}{\textit{DVC \citep{lu2019dvc} as Baseline} } & & & \\
\citet{li2016lambda}$^1$ & 20.21 & 17.13 & 13.71 & 10.32 & 16.69 \\
\citet{li2022rate}$^1$ & -6.80 & -2.96 & 0.48 & -6.85 & -4.12 \\
OEU \citep{lu2020content}$^2$ & -13.57 & -11.29 & -18.97 & -12.43 & -13.78 \\
BAO \citep{bao2022}$^2$ & -28.55 & -26.82 & -25.37 & -32.54 & -27.68 \\ 
Proposed  & -32.10 & -31.71 & -35.86 & -32.93 & -30.92 \\
\midrule
\multicolumn{3}{@{}l@{}}{\textit{DCVC \citep{li2021deep} as Baseline} } & & &  \\
OEU \citep{lu2020content}$^2$ & -10.75 & -14.34 & -16.30 & -7.15 & -16.07 \\
BAO \citep{bao2022}$^2$ & -20.59 & -19.69 & -20.60 & -23.33 & -25.22 \\
Proposed  & -32.89 & -33.10 & -32.01 & -36.88 & -39.66 \\\bottomrule
\end{tabular}
\end{small}
\captionof{table}{The BD-BR of our approach compared with others. $^1$ comes from \citet{li2022rate}. $^2$ comes from \citet{bao2022}.}
\label{tab:bdbr}
\end{minipage}
\hfill
\begin{minipage}[t]{0.295\textwidth}
\vspace{0pt}
\centering
\includegraphics[width=\textwidth]{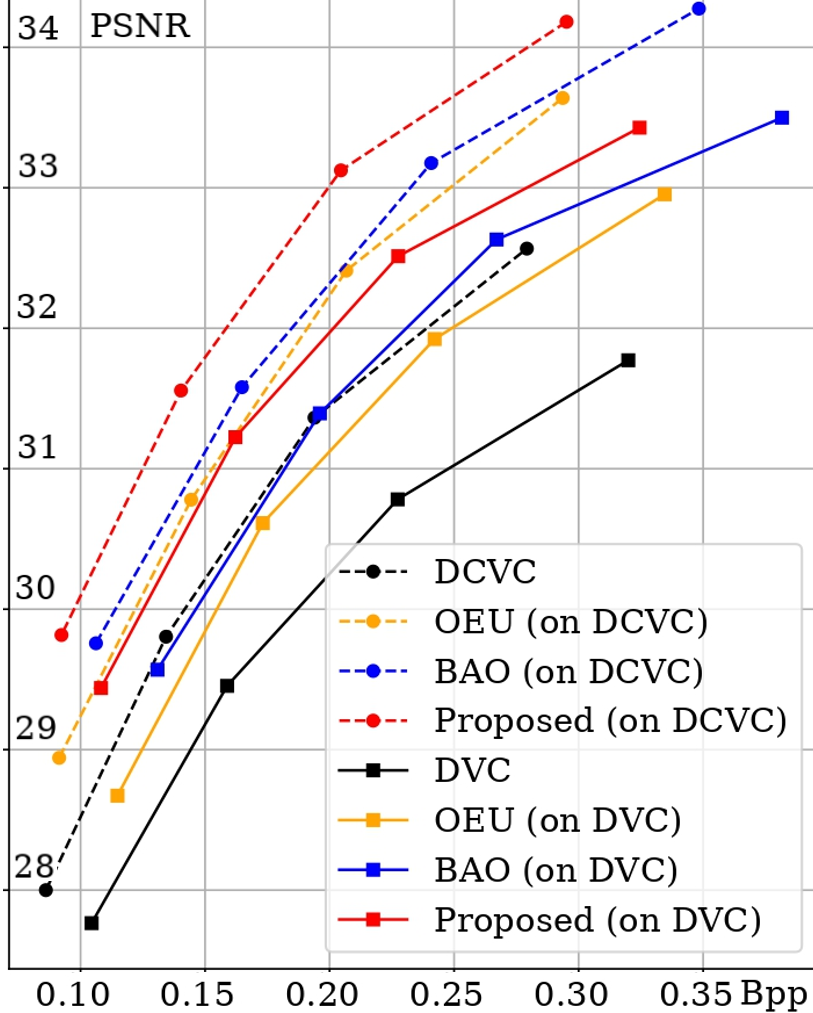}
\captionof{figure}{The R-D curve on HEVC Class D.}
\label{fig:bdbrd}
\end{minipage}
\end{minipage}

Other than R-D performance, the bitrate error of our approach is also significantly smaller than BAO \citep{bao2022} (See Tab.~\ref{tab:bderr}). The bitrate error is measured as the relative bitrate difference before and after bit allocation. The smaller it is, the easier it is to achieve the desired bitrate accurately. For complexity, our approach only performs $920$ steps of gradient ascent per-frame, while BAO requires $2000$ steps.

See more quantitative results (BD-PSNR \& R-D curves) in Appendix.~\ref{app:quant}.

\subsection{Ablation Study, Analysis \& Qualitative Results }
Tab.~\ref{tab:abl} shows that for BAO \citep{bao2022}, jointly optimizing $\bm{w}_{1:T},\bm{y}_{1:T}$ performs worse than optimizing $\bm{y}_{1:T}$ or $\bm{w}_{1:T}$ alone. This counter-intuitive phenomena comes from its incorrect estimation of gradient signal. For the proposed approach that corrects this, jointly optimizing $\bm{w}_{1:T},\bm{y}_{1:T}$ performs better than optimizing $\bm{y}_{1:T}$ or $\bm{w}_{1:T}$ alone, which is aligned with our intuition. 

\begin{minipage}{\textwidth}
\vspace{7pt}
\begin{minipage}[t]{0.68\textwidth}
\vspace{0pt}
\centering
\begin{small}
\begin{tabular}{@{}llllll@{}}
\toprule
 & \multicolumn{5}{c}{Bitrate-Error (\%) $\downarrow$} \\ \cmidrule(rr){2-6}
Method & Class B & Class C & Class D & Class E & UVG  \\ \midrule
\multicolumn{3}{@{}l@{}}{\textit{DVC \citep{lu2019dvc} as Baseline} } & & & \\
BAO \citep{bao2022}$^2$ & 8.41 & 12.86 & 21.39 & 5.94 & 3.73 \\ 
Proposed  & 3.16 & 4.27 & 1.81 & 6.14 & 1.73 \\
\midrule
\multicolumn{3}{@{}l@{}}{\textit{DCVC \citep{li2021deep} as Baseline} } & & &  \\
BAO \citep{bao2022}$^2$ & 25.67 & 23.90 & 23.74 & 24.88 & 21.86 \\
Proposed  & 4.27 & 7.29 & 5.73 & 8.03 & 3.06 \\\bottomrule
\end{tabular}
\end{small}
\captionof{table}{The bitrate error of our approach compared with BAO.}
\label{tab:bderr}
\end{minipage}
\hfill
\begin{minipage}[t]{0.30\textwidth}
\vspace{0pt}
\centering
\begin{small}
\begin{tabular}{@{}ll@{}}
\toprule
Method         & BD-BR (\%) $\downarrow$\\ \midrule
BAO ($\bm{y}$)        & -25.37 \\
BAO ($\bm{w}$)        & -22.24 \\
BAO ($\bm{y},\bm{w}$) & -14.76 \\
Proposed ($\bm{y}$)   & -32.60    \\
Proposed ($\bm{w}$)   & -31.56     \\
Proposed ($\bm{y},\bm{w}$) &  -35.86 \\ \bottomrule
\end{tabular}
\end{small}
\captionof{table}{Ablation study with HEVC Class D and DVC \citep{lu2019dvc}.}
\label{tab:abl}
\end{minipage}
\end{minipage}

\begin{figure}[h]
\centering
 \includegraphics[width=\linewidth]{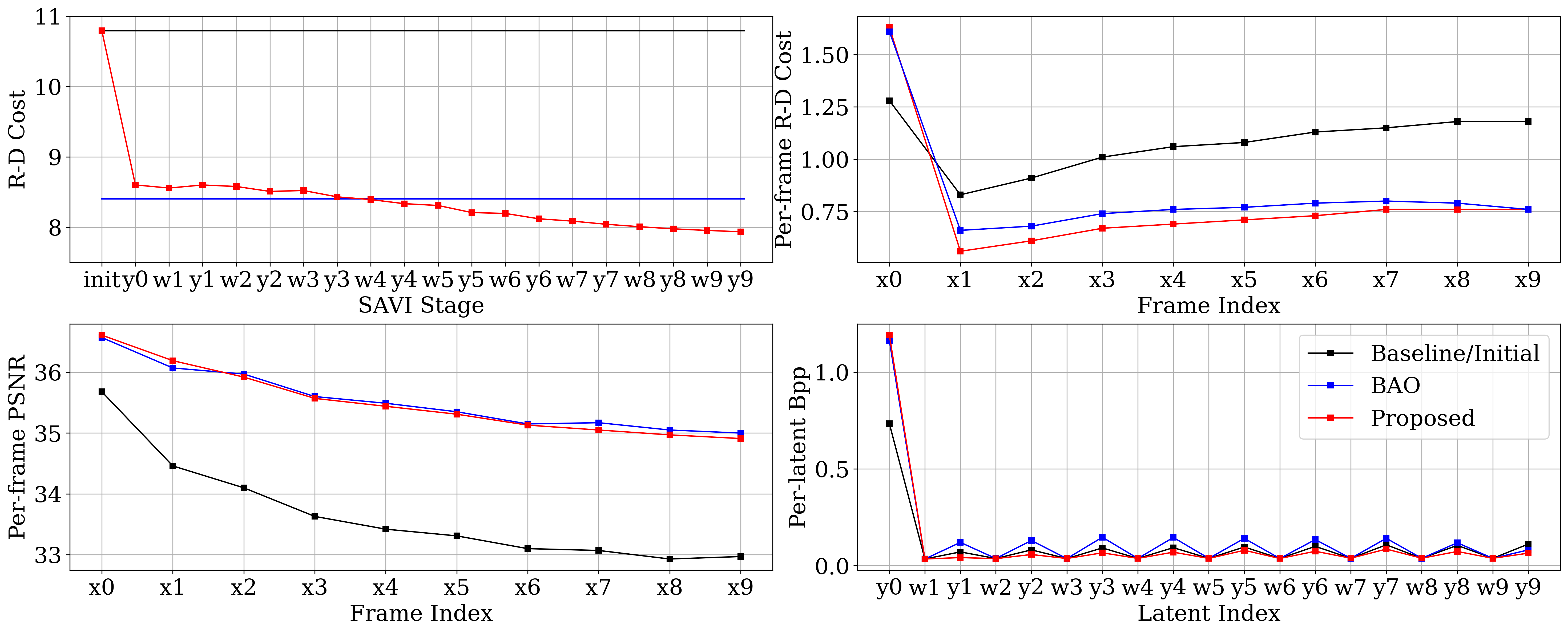}
 \caption{\textit{top-left}. R-D cost vs. SAVI stage. \textit{top-right}. R-D cost vs. frame index. \textit{bottom-left}. PSNR vs. frame index. \textit{bottom-right}. bpp vs. latent index. See enlarged-version in Appendix.~\ref{app:ana}.}
 \label{fig:anamain}
\end{figure}

To better understand why our method works, we present the R-D cost, distortion and rate versus frame/latent index for different methods in Fig.~\ref{fig:anamain}: \textit{top-left} shows that the R-D cost of our approach consistently decreases according to SAVI stage. Moreover, it outperforms BAO after $4^{th}$ frame; \textit{top-right} shows that for each frame the R-D cost of our method is lower than BAO; \textit{bottom-left} shows that the distortion part of R-D cost of our approach is approximately the same as BAO. While \textit{bottom-right} shows that the advantage of our approach over BAO lies in the bitrate. More specifically, BAO increases the bitrate of $\bm{y}_i$s after SAVI, while our correction decreases it.

See more analysis in Appendix.~\ref{app:ana} and qualitative results in Appendix.~\ref{app:qual}.

\section{Discussion \& Conclusion}
Despite our correction is already more efficient than original BAO \citep{bao2022}, its encoding speed remains far from real-time. Thus, it is limited to scenarios where R-D performance matters much more than encoding time (e.g. video on demand). See more discussion in Appendix.~\ref{app:disc}.

To conclude, we show that a previous bit allocation method for NVC is sub-optimal as it abuses SAVI on non-factorized latent. Then, we propose the correct SAVI on general non-factorized latent by back-propagating through gradient ascent, and we further propose a feasible approximation to make it practical for bit allocation. Experimental results show that our correction significantly improves the R-D performance. 


\subsubsection*{Ethics Statement}
Improving the R-D performance of NVC has positive social value, in terms of reducing carbon emission by saving the resources required to transfer and store videos. Moreover, unlike traditional codecs such as H.266 \citep{bross2021developments}, neural video codec does not require dedicated hardware. Instead, it can be deployed with general neural accelerators. Improving the R-D performance of NVC prompts the practical deployment of video codecs that are independent of dedicated hardware, and lowers the hardware-barrier of playing multi-media contents.

\subsubsection*{Reproducibility Statement}
For theoretical results, both of the two theorems are followed by proof in Appendix.~\ref{app:pf}. For a relatively complicated novel algorithm (Alg.~\ref{alg:solve-dag}), we provide an illustration of the step by step execution procedure in Appendix.~\ref{app:eg}. For experiment, both of the two datasets are publicly accessible. In Appendix.~\ref{app:impl}, we provide more implementation details including all the hyper-parameters. Moreover, we provide our source code for reproducing the empirical results in supplementary material.
\bibliography{iclr2023_conference}
\bibliographystyle{iclr2023_conference}

\newpage

\appendix

\section{Appendix}
\subsection{Proof of Thm~\ref{th:2l} and Thm~\ref{th:dag}}
\label{app:pf}
\begin{customthm}{1}
\label{th:2lcp}
After the procedure \textup{grad-2-level($\bm{x},\bm{w}^k$)} of Alg.~\ref{alg:solve-2} executes, we have the return value $d \mathcal{L}(\bm{w}^k,\bm{y}^K)/d\bm{w}^k=\overleftarrow{\bm{w}}$.
\end{customthm}
\begin{proof}
This proof extends the proof of Thm.~1 in \citet{domke2012generic}, and it also serves as a formal justification of Alg.~1 in \citet{kim2018semi}. Note that our paper and \citet{kim2018semi} are subtly different from \citet{samuel2009learning,domke2012generic} as our high level parameter $\bm{w}$ not only generate low level parameter $\bm{y}$, but also directly contributes to optimization target (See Fig.~\ref{fig:cg}).

\begin{figure}[htb]
\centering
    \includegraphics[width=0.35 \linewidth]{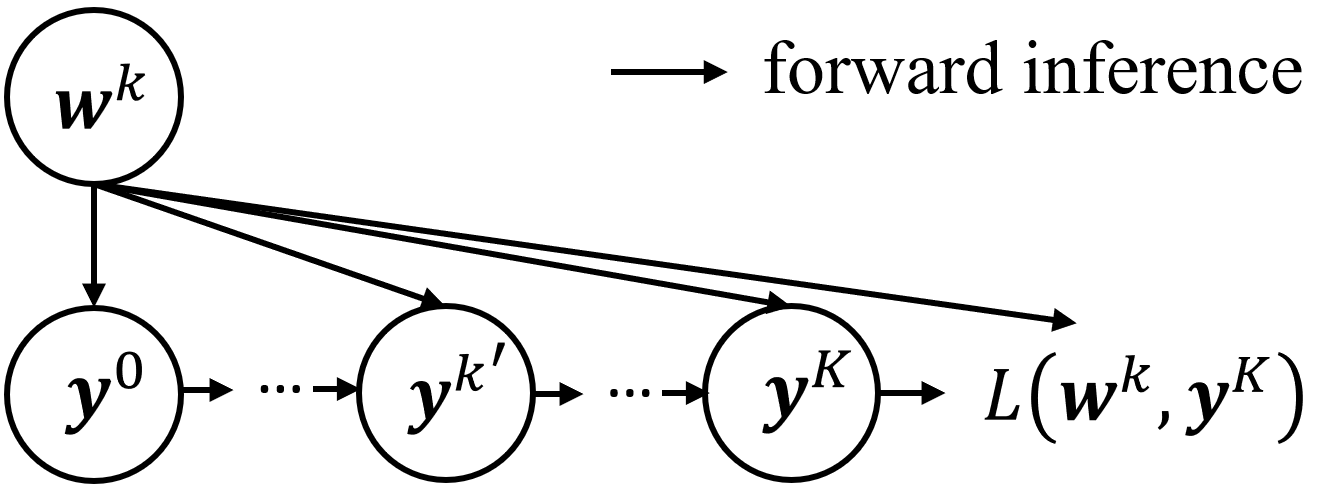}
    \caption{The computational graph corresponding to Eq.~\ref{eq:wlgd_gd}}
    \label{fig:cg}
\end{figure}
As the computational graph in Fig.~\ref{fig:cg} shows, we can expand $d \mathcal{L}(\bm{w}^k,\bm{y}^{K})/d\bm{w}^k$ as Eq.~\ref{eq:wlgd_gd}, with each term solved in Eq.~\ref{eq:yw_gd} and Eq.~\ref{eq:yKyk_gd}.
\begin{align}
    \frac{d \mathcal{L}(\bm{w}^k,\bm{y}^{K})}{d\bm{w}^k}=\underbrace{\frac{\partial \mathcal{L}(\bm{w}^k,\bm{y}^{K})}{\partial \bm{w}^k}}_{\textrm{known}}+\sum_{k'=0}^{K}\underbrace{\frac{\partial \bm{y}^{k'}}{\partial\bm{w}^k}}_{\textrm{Eq.~}\ref{eq:yw_gd}}\underbrace{\frac{d\mathcal{L}(\bm{w}^k,\bm{y}^K)}{d\bm{y}^{k'}}}_{\textrm{Eq.~}\ref{eq:yKyk_gd}}
    \label{eq:wlgd_gd}
\end{align}
To solve Eq.~\ref{eq:wlgd_gd}, we first note that $\partial\mathcal{L}(\bm{w}^k,\bm{y}^K)/\partial\bm{w}^k,d\mathcal{L}(\bm{w}^k,\bm{y}^K)/d\bm{y}^K,\partial\bm{y}^0/\partial\bm{w}^k$ is naturally known. Then, by taking partial derivative of the update rule of gradient ascent $\bm{y}^{k'+1}\leftarrow \bm{y}^{k'}+\alpha d\mathcal{L}(\bm{w}^k,\bm{y}^{k'})/d\bm{y}^{k'}$ with regard to $\bm{w}^k,\bm{y}^{k'}$, we have Eq.~\ref{eq:yy_gd} and Eq.~\ref{eq:yw_gd}. Note that Eq.~\ref{eq:yw_gd} is the partial derivative $\partial \bm{y}^{k'+1}/\partial\bm{w}^k$ instead of total derivative $d\bm{y}^{k'+1}/d\bm{w}^k=(\partial \bm{y}^{k'+1}/\partial\bm{y}^{k'}) (d\bm{y}^{k'}/d\bm{w}^k)+\partial \bm{y}^{k'+1}/\partial\bm{w}^k$.
\begin{align}
    \frac{\partial \bm{y}^{k'+1}}{\partial\bm{y}^{k'}}&=I+\alpha\frac{\partial^2\mathcal{L}(\bm{w}^k,\bm{y}^{k'})}{\partial\bm{y}^{k'}\partial\bm{y}^{k'}}\label{eq:yy_gd}\\
    \frac{\partial \bm{y}^{k'+1}}{\partial\bm{w}^k}&=\alpha\frac{\partial^2\mathcal{L}(\bm{w}^k,\bm{y}^{k'})}{\partial\bm{w}^k\partial\bm{y}^{k'}}
    \label{eq:yw_gd}
\end{align}
And those second order terms can either be directly evaluated or approximated via finite difference as Eq.~\ref{eq:hes}. As Eq.~\ref{eq:yw_gd} already solves the first term on the right hand side of Eq.~\ref{eq:wlgd_gd}, the remaining issue is $d\mathcal{L}(\bm{w}^k,\bm{y}^K)/d\bm{y}^{k'}$. To solve this term, we expand it recursively as Eq.~\ref{eq:yKyk_gd} and take Eq.~\ref{eq:yy_gd} into it.
\begin{align}
    \frac{d\mathcal{L}(\bm{w}^k,\bm{y}^K)}{d\bm{y}^{k'}}=\frac{\partial \bm{y}^{k'+1}}{\partial\bm{y}^{k'}}\frac{d\mathcal{L}(\bm{w}^k,\bm{y}^K)}{d\bm{y}^{k'+1}}
    \label{eq:yKyk_gd}
\end{align}
And the above solving process can be described by the procedure grad-2-level($\bm{x},\bm{w}^k$) of Alg.~\ref{alg:solve-2}. Specifically, the iterative update of $\overleftarrow{\bm{y}^{k'+1}}$ in line 15 corresponds to recursively expanding Eq.~\ref{eq:yKyk_gd} with  Eq.~\ref{eq:yy_gd}, and the iterative update of $\overleftarrow{\bm{w}}$ in line 14 corresponds to recursively expanding Eq.~\ref{eq:wlgd_gd} with Eq.~\ref{eq:yw_gd} and Eq.~\ref{eq:yKyk_gd}. Upon the return of grad-2-level($\bm{x},\bm{w}^k$) of Alg.~\ref{alg:solve-2}, we have $\overleftarrow{\bm{w}}=d \mathcal{L}(\bm{w}^k,\bm{y}^{K})/d\bm{w}^k$.

The complexity of the Hessian-vector product in line 14 and 15 of Alg.~\ref{alg:solve-2} may be reduced using finite difference following \citep{domke2012generic} as Eq.~\ref{eq:hes}.
\begin{align}
    \frac{\partial^2 \mathcal{L}(\bm{w}^{k},\bm{y}^{k'})}{\partial \bm{w}^{k}\partial\bm{y}^{k'}}\bm{v}&=\lim_{r\rightarrow 0}\frac{1}{r}(\frac{d \mathcal{L}(\bm{w}^{k},\bm{y}^{k'}+r\bm{v})}{d \bm{w}^{k}}-\frac{d \mathcal{L}(\bm{w}^{k},\bm{y}^{k'})}{d \bm{w}^{k}})\notag\\
    \frac{\partial^2 \mathcal{L}(\bm{w}^{k},\bm{y}^{k'})}{\partial \bm{y}^{k'}\partial\bm{y}^{k'}}\bm{v}&=\lim_{r\rightarrow 0}\frac{1}{r}(\frac{d \mathcal{L}(\bm{w}^{k},\bm{y}^{k'}+r\bm{v})}{d \bm{y}^{k'}}-\frac{d \mathcal{L}(\bm{w}^{k},\bm{y}^{k'})}{d \bm{y}^{k'}})\label{eq:hes}
\end{align}
\end{proof}
\begin{customthm}{2}
\label{th:dagcp}
After the procedure \textup{grad-dag($\bm{x},\bm{y}_0^{k_0},...,\bm{y}_i^{k_i}$)} in Alg.~\ref{alg:solve-dag} executes, we have the return value $d\mathcal{L}(\bm{y}_0^{k_0},...,\bm{y}_i^{k_i},\bm{y}_{>i}^K)/d \bm{y}_i^{k_i}=\overleftarrow{\bm{y}_i}$.
\end{customthm}
\begin{proof}
Consider computing the target gradient with DAG $\mathcal{G}$. The $\bm{y}_i^k$'s gradient is composed of its own contribution to $\mathcal{L}$ in addition to the gradient from its children $\bm{y}_j\in\mathcal{C}(\bm{y}_i)$. Further, as we are considering the optimized children $\bm{y}_j^K$, we expand the children node $\bm{y}_j$ as Fig.~\ref{fig:cg}. Then, we have:
\begin{align}
    \frac{d\mathcal{L}(\bm{y}_0^{k_0},...,\bm{y}_i^{k_i},\bm{y}_{>i}^K)}{d \bm{y}_i^{k_i}}=\underbrace{\frac{\partial \mathcal{L}(\bm{y}_0^{k_0},...,\bm{y}_i^{k_i},\bm{y}_{>i}^K)}{\partial \bm{y}_i^{k_i}}}_{\textrm{known}} + \sum_{\bm{y}_j\in\mathcal{C}(\bm{y}_i)}(\sum_{k_j=0}^{K}\underbrace{\frac{\partial \bm{y}_j^{k_j}}{\partial\bm{y}_i^{k_i}}}_{\textrm{Eq.~}\ref{eq:yw_gd}}\underbrace{\frac{d\mathcal{L}(\bm{y}_0^{k_0},...,\bm{y}_{j-1}^{k_{j-1}},\bm{y}_{\ge j}^K)}{d\bm{y}_j^{k_j}}}_{\textrm{Eq.~}\ref{eq:yKyk_gd}})\label{eq:graddag}
\end{align}
The first term on the right-hand side of Eq.~\ref{eq:graddag} can be trivially evaluated. The $\partial\bm{y}_j^{k_j}/\partial\bm{y}_i^{k_i}$ can be evaluated as Eq.~\ref{eq:yw_gd}. And the $ d\mathcal{L}(\bm{y}_0^{k_0},...,\bm{y}_{j-1}^{k_{j-1}},\bm{y}_{\ge j}^K)/d\bm{y}_j^{k_j}$ can be iteratively expanded as Eq.~\ref{eq:yKyk_gd}. We highlight several key differences between Alg.~\ref{alg:solve-dag} and Alg.~\ref{alg:solve-2} which are reflected in the implementation of Alg.~\ref{alg:solve-dag}:
\begin{itemize}
    \item The gradient evaluation of current node $\bm{y}_i$ requires gradient of its plural direct children $\bm{y}_j\in\mathcal{C}(\bm{y}_i)$, instead of the single child in 2-level case. The children traversal part of Eq.~\ref{eq:yKyk_gd} corresponds to the two extra for loop in line 8 and 14 of Alg.~\ref{alg:solve-dag}. 
    \item The gradient ascent update of child latent parameter $\bm{y}_j^{k_j+1}\leftarrow \bm{y}_j^{k_j}+\alpha d\mathcal{L}(\bm{y}_0^{k_0},...,\bm{y}_j^{k_j},\bm{y}_{>j}^K)/d\bm{y}_{j}^{k_j}$ can be conducted trivially only if $\mathcal{C}(\bm{y}_j)$ is empty, otherwise the gradient has to be evaluated recursively using Eq.~\ref{eq:graddag}. And this part corresponds to the recursive call in line 11 of Alg.~\ref{alg:solve-dag}.
\end{itemize}
And the other part of Alg.~\ref{alg:solve-dag} is the same as Alg.~\ref{alg:solve-2}. So the rest of the proof follows Thm.~\ref{th:2l}. Similarly, the Hessian-vector product in line 17 and 18 of Alg.~\ref{alg:solve-dag} may be approximated as Eq.~\ref{eq:hes}. However, this does not save Alg.~\ref{alg:solve-dag} from an overall complexity of $\Theta(K^N)$.
\end{proof}
\subsection{The Complete Formula for Sec.~\ref{sec:itde} and Sec.~\ref{sec:ipde}}
\label{app:cf}
In this section, we provide the complete formula on $\bm{y}_i$ related gradient for Sec.~\ref{sec:itde} and Sec.~\ref{sec:ipde}. Specifically, Eq.~\ref{eq:incyg} is paired with Eq.~\ref{eq:incwg}, and Eq.~\ref{eq:incygy} is paired with Eq.~\ref{eq:incygw}.
\begin{align}
\frac{d \mathcal{L}(\bm{w}_{1:T},\bm{y}_{1:T})}{d \bm{y}_i}=&\sum_{j=i}^{T}\frac{d \mathcal{L}_j(\bm{w}_{1:j},\bm{y}_{1:j})}{d \bm{y}_i}\notag\\
\frac{d \mathcal{L}_j(\bm{w}_{1:j},\bm{y}_{1:j})}{d \bm{y}_i}=&\underbrace{\sum_{l=i+1}^{j}(\frac{\partial \bm{y}_{l}}{\partial\bm{y}_i}\frac{d \mathcal{L}_j(\bm{w}_{1:j},\bm{y}_{1:j})}{d \bm{y}_l}+\frac{\partial \bm{w}_l}{\partial\bm{y}_i}\frac{d \mathcal{L}_j(\bm{w}_{1:j},\bm{y}_{1:j})}{d \bm{w}_l})}_{\textrm{ignored by BAO}}+\underbrace{\frac{\partial \mathcal{L}_j(\bm{w}_{1:j},\bm{y}_{1:j})}{\partial \bm{y}_i}}_{\textrm{considered by BAO}}\label{eq:incyg}
\end{align}
\begin{align}
\bm{y}^{k'_i+1}_i \leftarrow \bm{y}^{k'_i} + \alpha\frac{d \mathcal{L}(\bm{w}_1^{k_1},...,\bm{w}_i^{k_i},\bm{w}^K_{>i},\bm{y}_1^{k'_1},...,\bm{y}_i^{k'_i},\bm{y}^K_{>i})}{d \bm{y}^{k'_i}_i},\notag\\\textrm{where }\bm{w}_{>i}^0,\bm{y}_{> i}^0=f(\bm{x},\bm{w}_1^{k_1},...,\bm{w}_i^{k_i},\bm{y}_1^{k'_1},...,\bm{y}_{i}^{k'_i}) \label{eq:incygy}
\end{align}
\subsection{An Example of Execution of Alg.~\ref{alg:solve-dag}}
\label{app:eg}
\begin{figure}[thb]
\centering
\includegraphics[width=\linewidth]{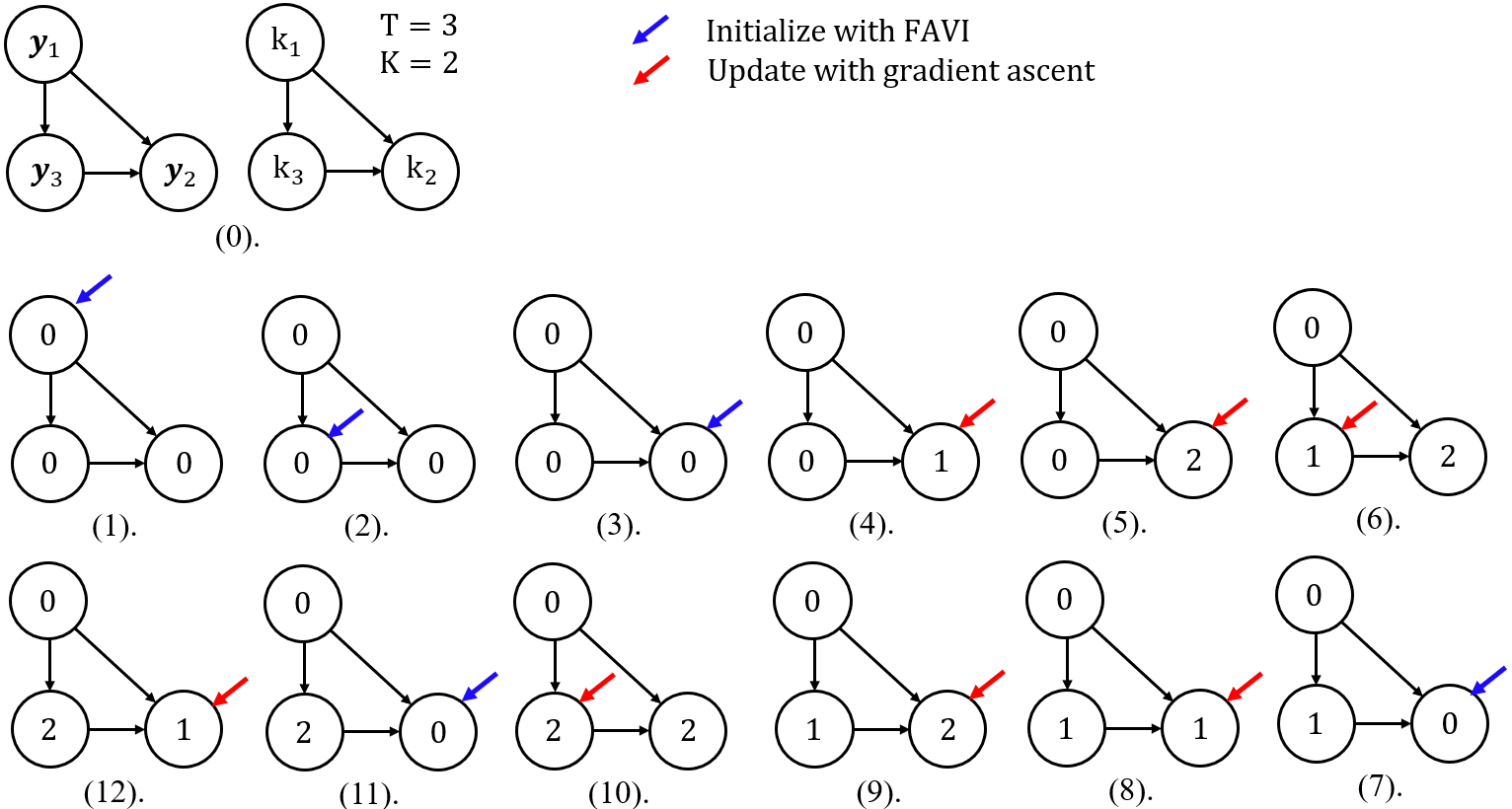}
\includegraphics[width=\linewidth]{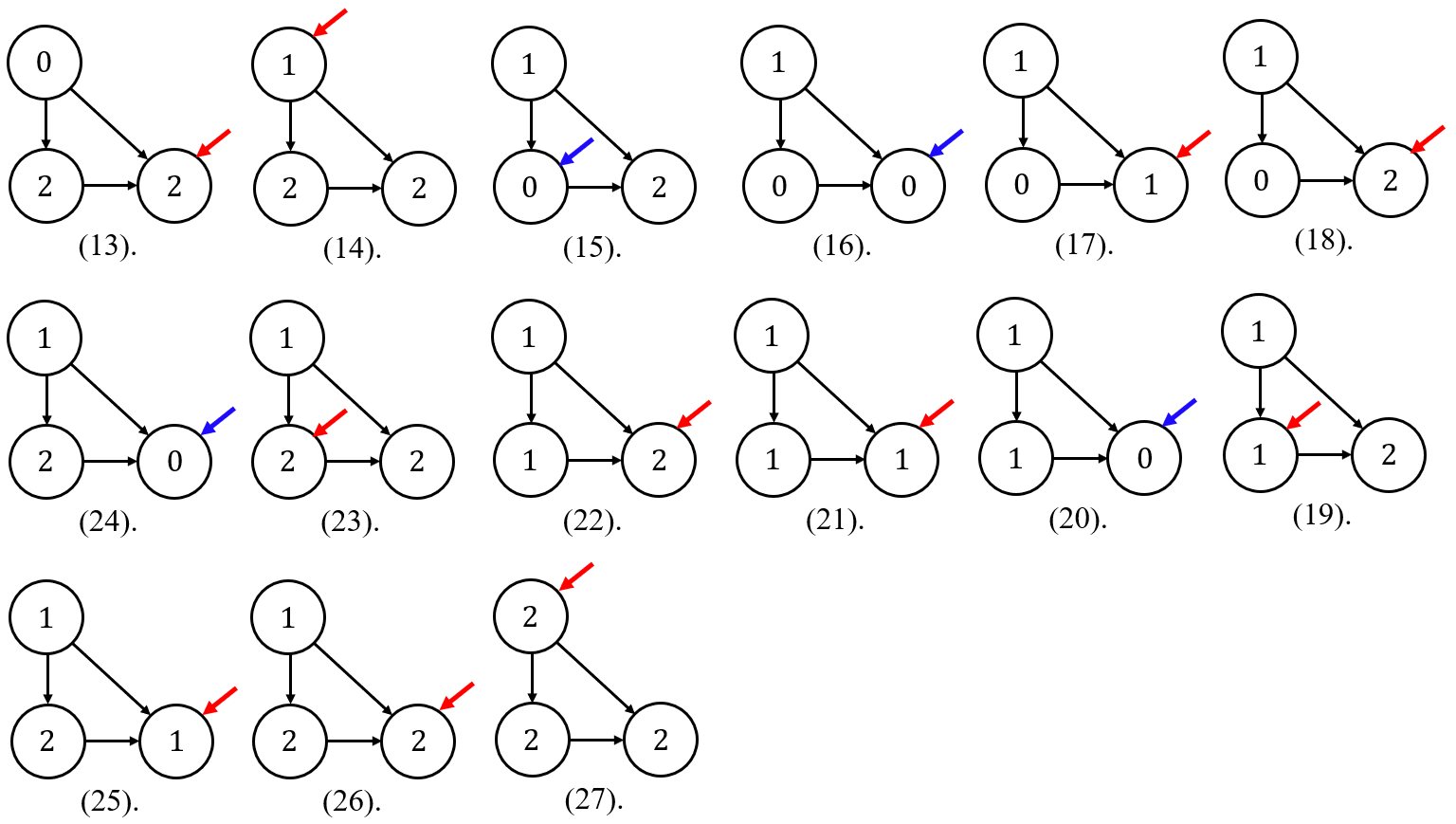}
\caption{(0). The example setup. (1).-(27). The execution procedure. The number in the circle indicates the gradient step $k_i$ of each node $\bm{y}_i$. The bold blue/red arrow indicates that the current node is under initialization/gradient ascent.}
\label{fig:eg}
\end{figure}

In this section, we provide an example of the full percedure of execution of Alg.~\ref{alg:solve-dag} in Fig.~\ref{fig:eg}. The setup is as Fig.~\ref{fig:eg}.(0): we have $N=3$ latent $\bm{y}_1,\bm{y}_2,\bm{y}_3$ and gradient ascent step $K=2$, connected by a DAG shown in the figure.
\subsection{More Implementation Details}
\label{app:impl}
In the main text, we use $\bm{y}_i$ as all the latent variable related to residual. In practice, it is divided into $\bm{y}_i,\bm{z}_i,\Delta^y_i$, which refer to the first level latent of residual, second level latent of residual and quantization step size of first level latent of residual respectively. In practice, as BAO \citep{bao2022}, all of the 3 parts are involved in SAVI jointly. We note that this is not a problem as they fully factorize. And for DVC \citep{lu2019dvc}, $\bm{w}_i$ indeed represent the latent of motion. As for DVC, the motion has only one level of latent. However for DCVC \citep{li2021deep}, $\bm{w}_i$ is divided into $\bm{w}_i,\bm{v}_i,\Delta^w_i$, which refer to the first level latent of motion, second level latent of motion and quantization step size of first level latent of motion respectively. Similar to $\bm{y}_i$, all of the 3 parts are involved in SAVI jointly, and this is not a problem as they fully factorize. 

Following BAO \citep{bao2022}, we set the target $\lambda_0=\{256,512,1024,2048\}$, which also follows the baselines \citep{lu2019dvc,li2021deep}. We adopt the official pre-train models for both of the baseline methods \citep{lu2019dvc,li2021deep}. We do not have a training dataset or implementation details for training amortized encoder / decoder as all the experiments are performed on official pre-trained models. For gradient ascent, we set $K=2000$ for the first I frame and $K=400$ for all other P frames. On average, the gradient ascent steps for each frame is $920$, which is smaller than $2000$ in BAO.

\subsection{More Quantitative Results}

\begin{figure}[thb]
\centering
    \begin{minipage}[t]{0.328\textwidth}
    \includegraphics[width=\textwidth]{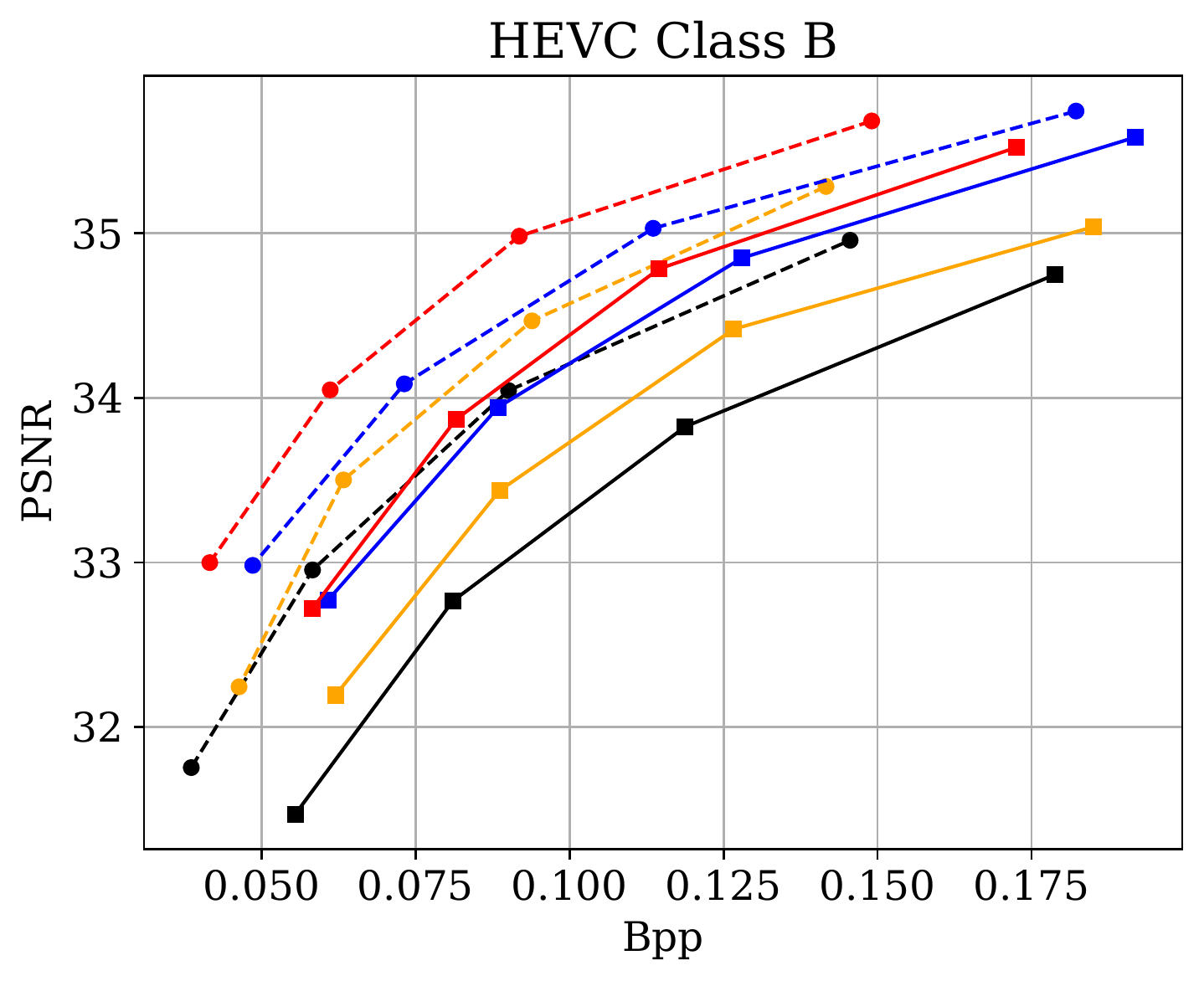}
    \end{minipage}%
    \hfill
    \begin{minipage}[t]{0.328\textwidth}
    \includegraphics[width=\textwidth]{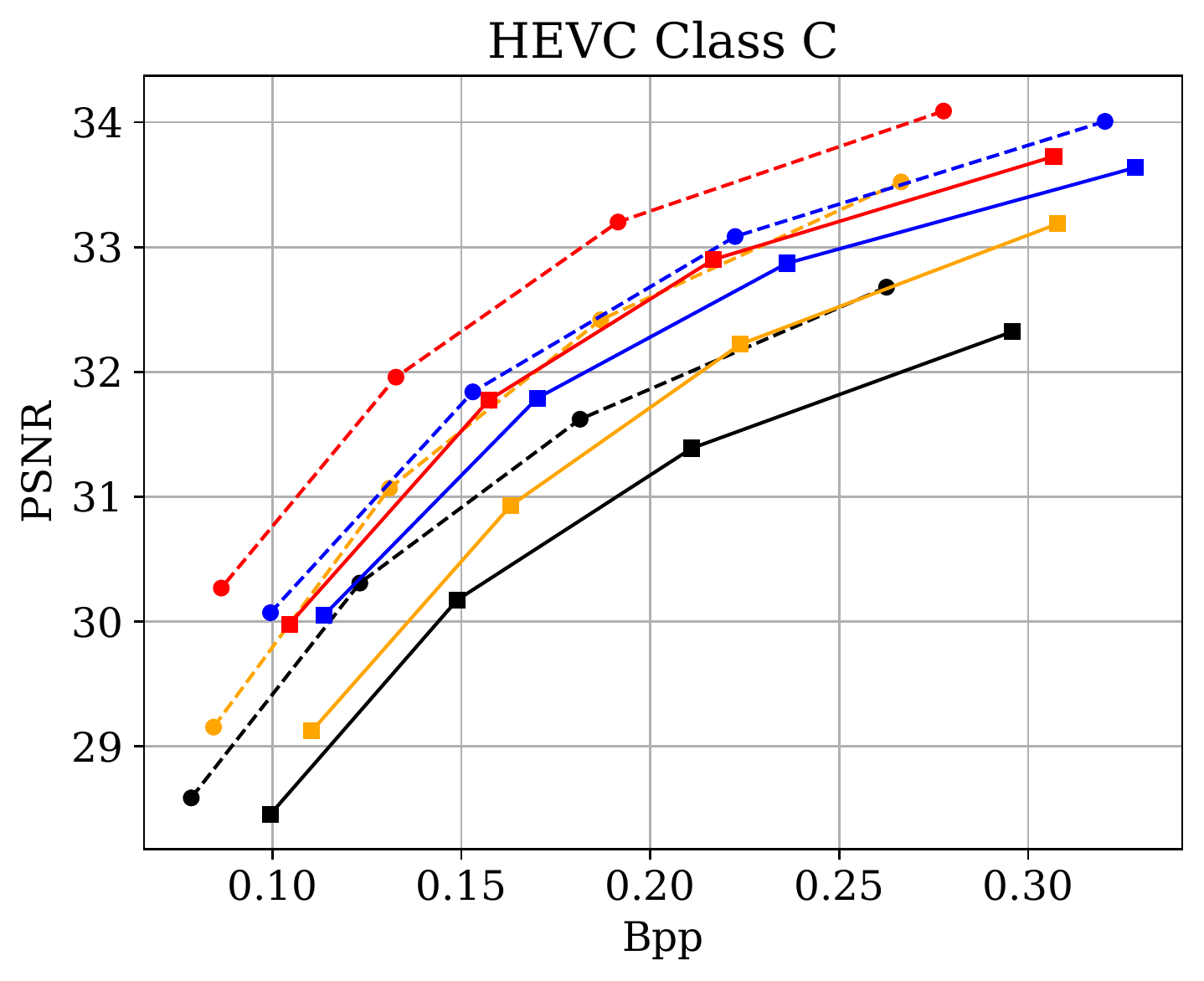}
    \end{minipage}%
    \hfill
    \begin{minipage}[t]{0.328\textwidth}
    \includegraphics[width=\textwidth]{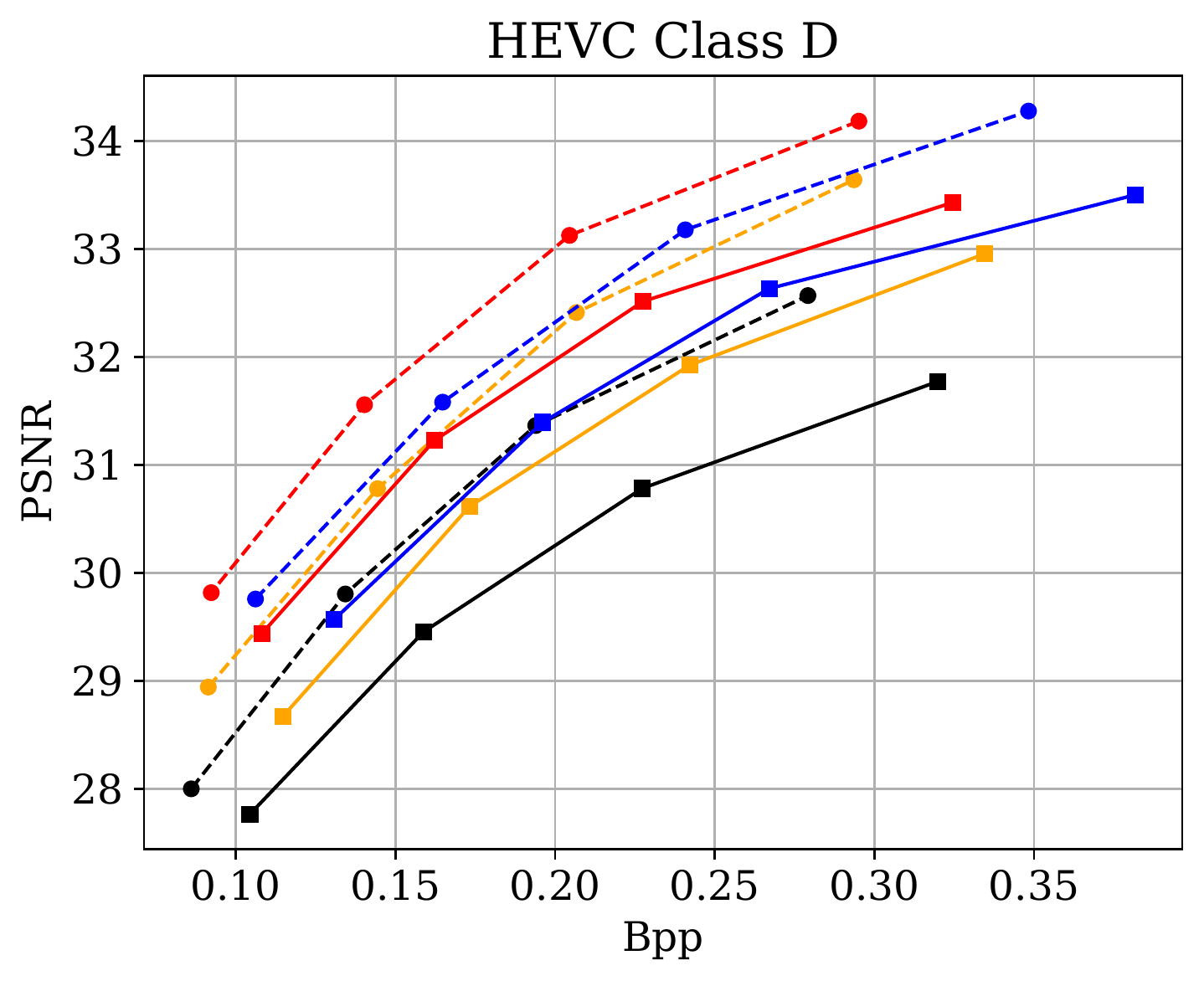}
    \end{minipage}%
    \vfill
    \vspace{0.2cm}
    \begin{minipage}[t]{0.328\textwidth}
    \includegraphics[width=\textwidth]{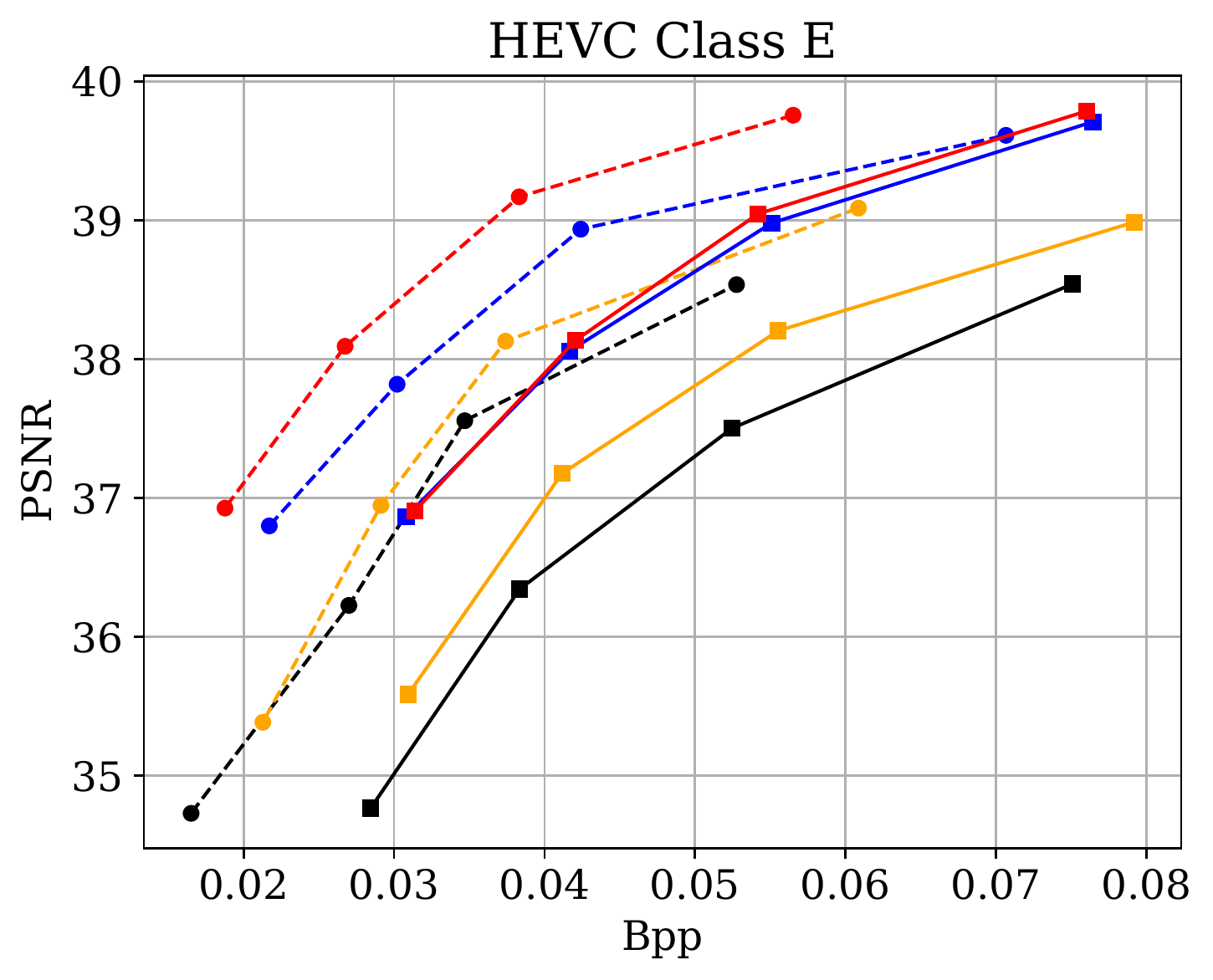}
    \end{minipage}%
    \hspace{0.002cm}
    \begin{minipage}[t]{0.328\textwidth}
    \includegraphics[width=\textwidth]{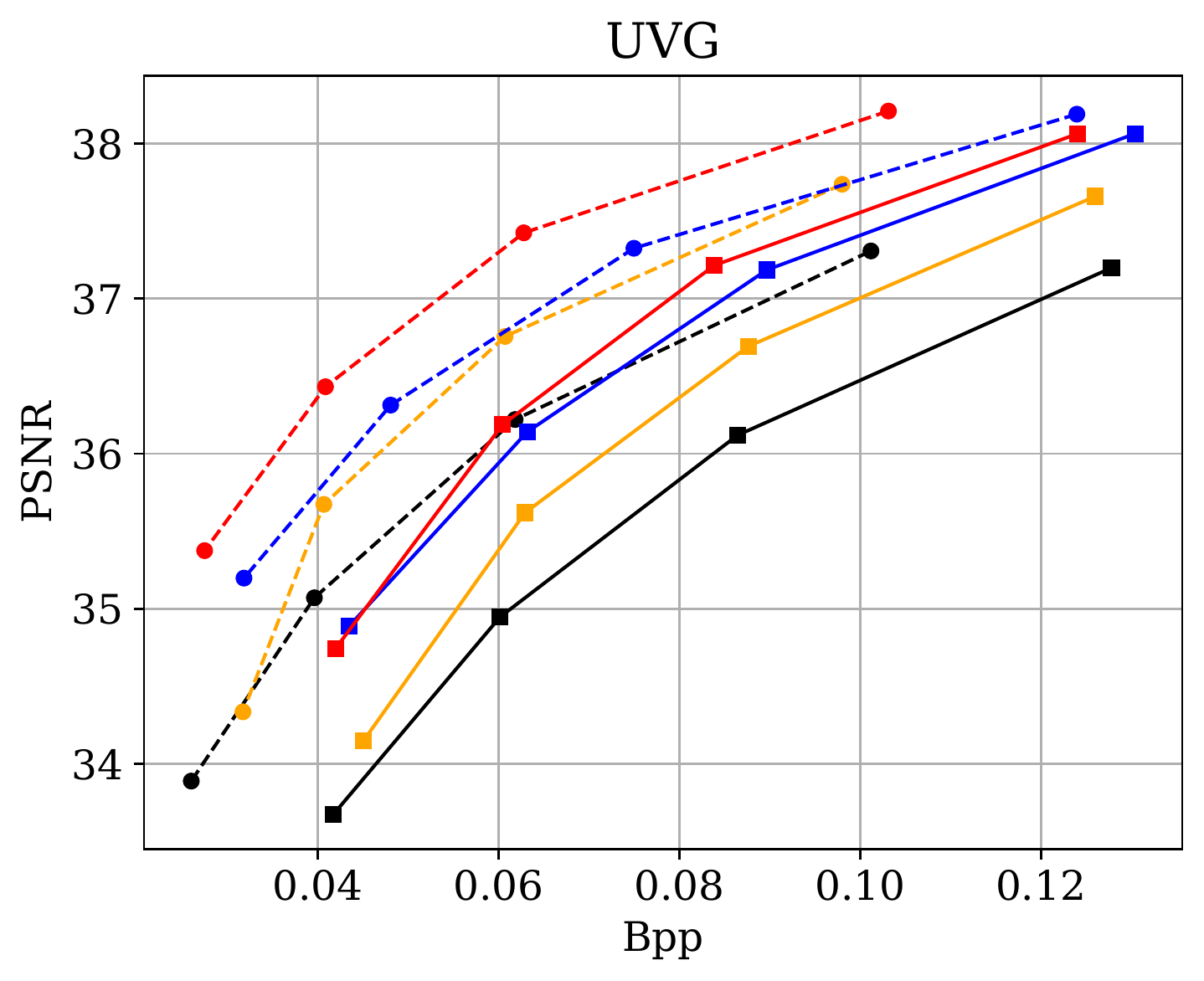}
    \end{minipage}%
    \hspace{0.1cm}
    \begin{minipage}[t]{0.31\textwidth}
    \includegraphics[width=\textwidth]{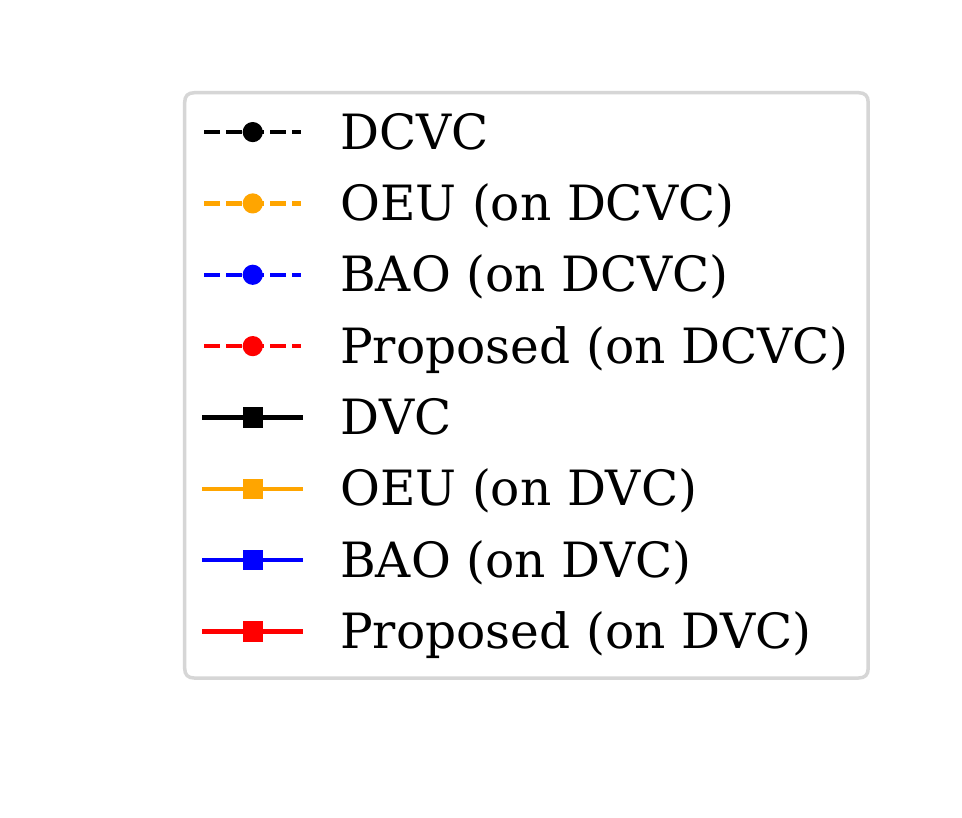}
    \end{minipage}%
    \caption{The R-D performance of our approach compared with baselines (w/o bit allocation) and other bit allocation approaches.}
    \label{fig:rd_all}
\end{figure}

In this section we present more quantitative results. In Tab.~\ref{tab:bdpsnr} we show the BD-PSNR of our proposed method and other methods as a supplementary to the BD-BR results (Tab.~\ref{tab:bdbr}). Furthermore, in Fig.~\ref{fig:rd_all}, we present R-D curve on all classes of HEVC CTC and UVG dataset as a supplementary to the HEVC Class D plot (Fig.~\ref{fig:bdbrd}).
\label{app:quant}
\begin{table*}[thb]
\centering
\begin{tabular}{@{}llllll@{}}
\toprule
 & \multicolumn{5}{c}{BD-PSNR (dB) $\uparrow$} \\ \cmidrule(rr){2-6}
Method & Cls B & Cls C & Cls D & Cls E & UVG \\ \midrule
\multicolumn{3}{@{}l@{}}{\textit{DVC \citep{lu2019dvc} as Baseline} } & & &  \\
\citet{li2016lambda}$^1$ &-0.54&-&-&-0.32&-0.47 \\
\citet{li2022rate}$^1$ &0.19&-&-&0.28 & 0.14 \\
OEU \citep{lu2020content} & 0.39 & 0.49 & 0.83 & 0.48 & 0.48 \\
BAO \citep{bao2022}$^2$ & 0.87 & 1.11 & 1.17 & 1.35 & 0.98 \\ 
Proposed  & 1.03 & 1.38  & 1.67 & 1.41 & 1.15 \\\midrule
\multicolumn{3}{@{}l@{}}{\textit{DCVC \citep{li2021deep} as Baseline} } & & &  \\
OEU \citep{lu2020content} & 0.30 & 0.58 & 0.74 & 0.29 & 0.50 \\
BAO \citep{bao2022}$^2$ & 0.52 & 0.76 & 0.96 & 0.96 & 0.69 \\ 
Proposed  & 0.91 & 1.37 & 1.55 & 1.58 & 1.18 \\ \bottomrule
\end{tabular}
\caption{The BD-PSNR of our approach compared with baselines (w/o bit allocation) and other bit allocation approaches. $^1$ comes from \citet{li2022rate}. $^2$ comes from \citep{bao2022}.}
\label{tab:bdpsnr}
\end{table*}

\subsection{More Analysis}
\label{app:ana}

\begin{figure}[thb]
\centering
 \includegraphics[width=\linewidth]{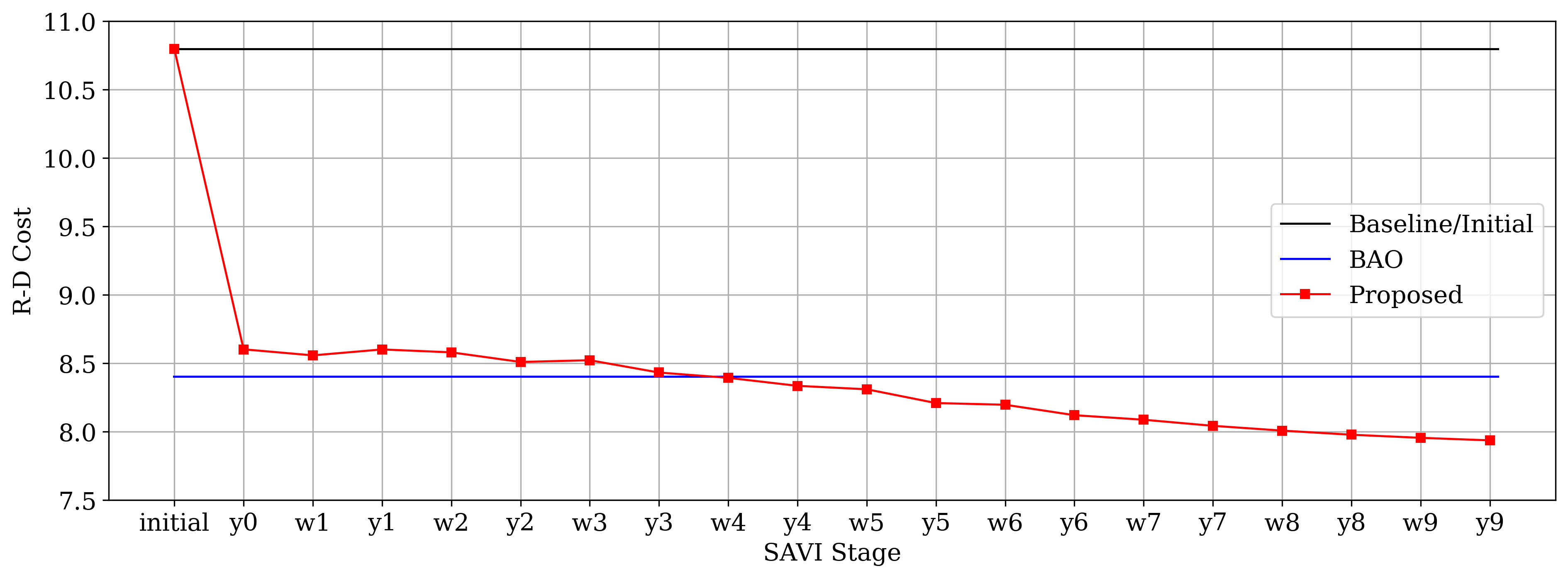}
 \caption{The R-D cost versus SAVI Stage. Experiment is conducted with DVC \citep{lu2019dvc} as baseline and \textit{BasketballPass} of HEVC Class D as data.}
 \label{fig:anal}
\includegraphics[width=\linewidth]{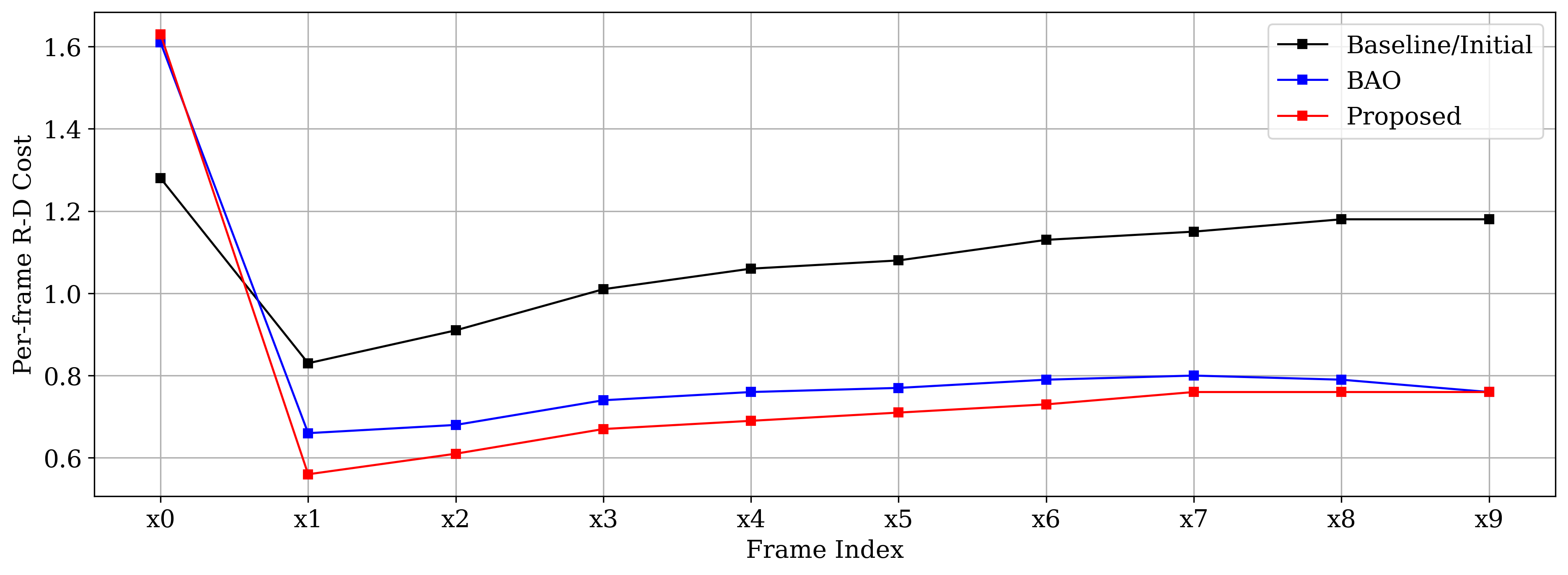}
\includegraphics[width=\linewidth]{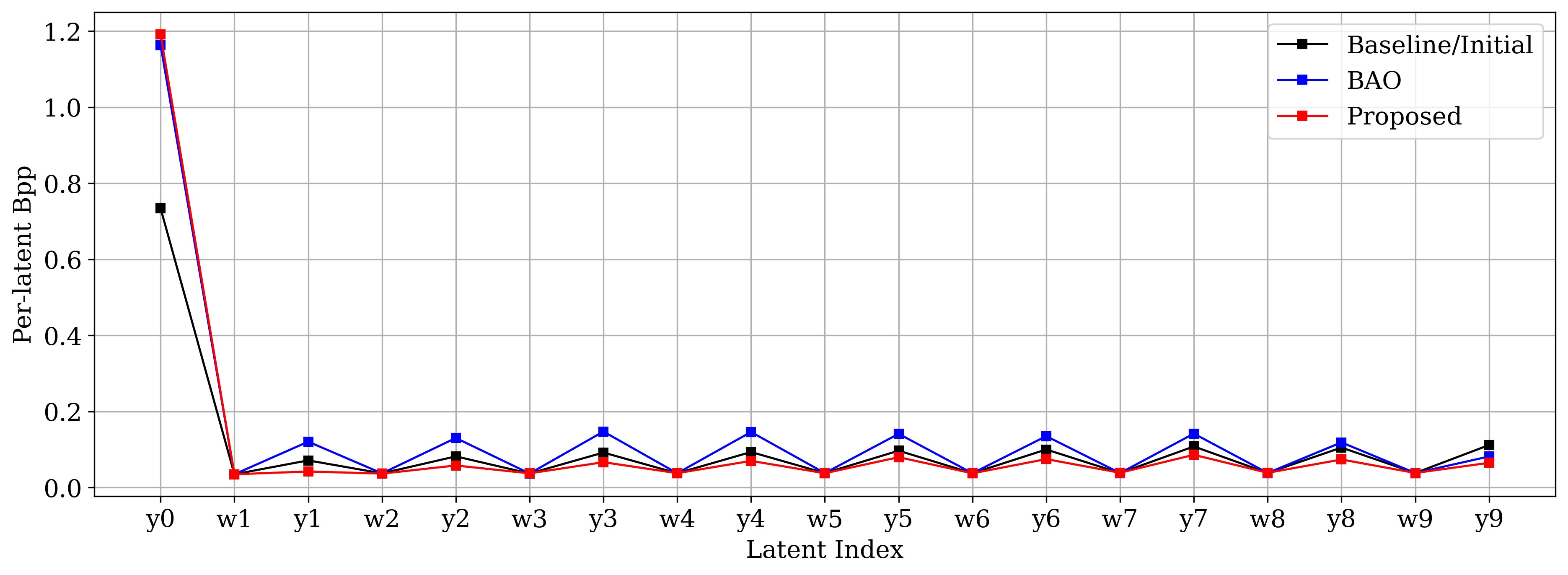}
\includegraphics[width=\linewidth]{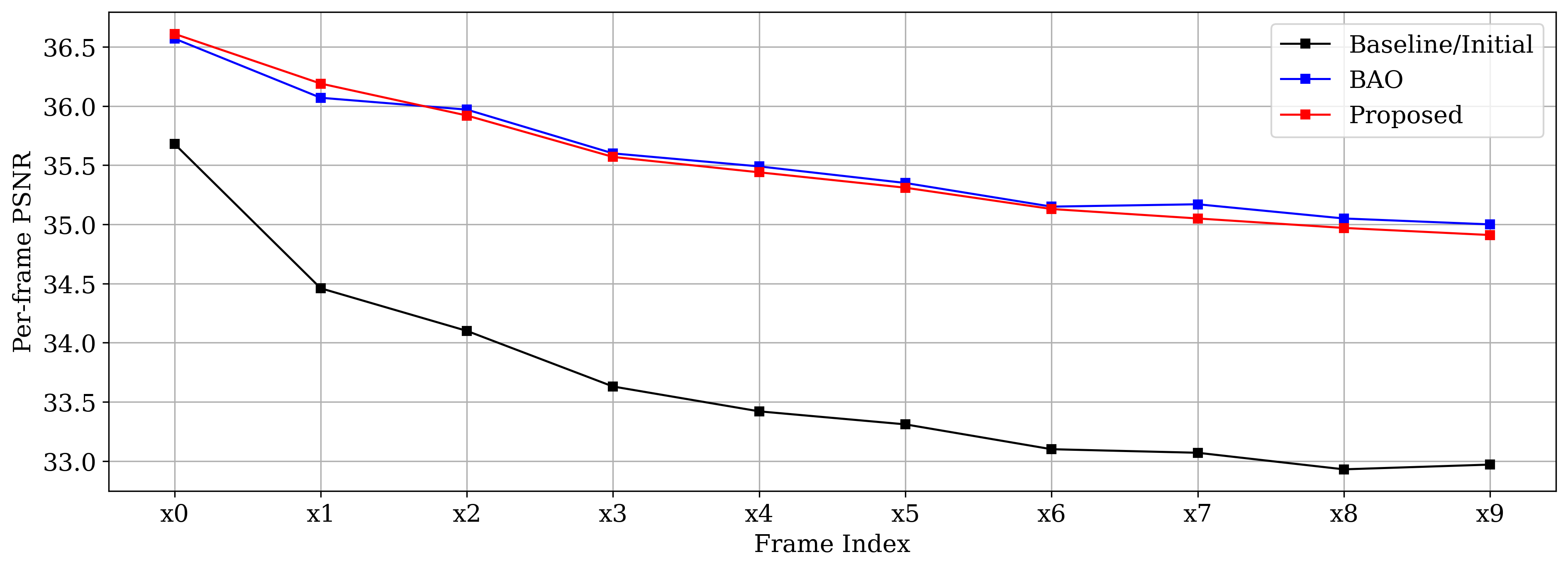}
\caption{The R-D cost, bpp and PSNR versus frame/latent index. Experiment is conducted with DVC \citep{lu2019dvc} as baseline and \textit{BasketballPass} of HEVC Class D as data.}
\label{fig:anal2}
\end{figure}

In this section, we extend the analysis on why the proposed approach works and what is the difference between the proposed approach and BAO \citep{bao2022}. In the approximate SAVI on DAG latent (Alg.~\ref{alg:solve-adag}), we solve SAVI approximately latent by latent in topological order. For bit allocation of NVC with $10$ frames, this topological order is $\bm{y}_0,\bm{w}_1,\bm{y}_1,...,\bm{w}_9,\bm{y}_9$, where $\bm{y}_0$ is the latent of I frame, $\bm{w}_i$ is the motion latent of $i^{th}$ P frame and $\bm{y}_i$ is the residual latent of $i^{th}$ P frame. In Fig.~\ref{fig:anal}, we show the relationship between R-D cost and the stage of approximate SAVI. We can see that the R-D cost reduces almost consistently with the growing of SAVI stage, which indicates that our approximate SAVI on DAG (Alg.~\ref{alg:solve-adag}) is successful. Specifically, despite our approach is inferior to BAO \citep{bao2022} upon the convergence of $\bm{y}_3$, it attains significant advantage over BAO after $\bm{y}_9$ converges.

In Fig.~\ref{fig:anal2}, we compare the distribution of R-D cost, PSNR and Bpp across frame and latent of the baseline DVC \citep{lu2019dvc}, BAO \cite{bao2022} and the proposed approach. For R-D cost, it is obvious that our proposed approach's R-D cost is lower than BAO and baseline, which indicates a better R-D performance. For bpp, it is interesting to observe that despite all three methods have similar bpp of motion related latent $\bm{w}_{1:T}$, the bpp of residual related latent $\bm{y}_{1:T}$ is quite different. Specifically, BAO increases the bpp of $\bm{y}_{1:T}$ compared with baseline, while our approach decreases the bpp of $\bm{y}_{1:T}$ compared with baseline. This explains why our approach has lower bitrate compared with BAO, and also explains why our approach has significantly less bitrate error. For the PSNR metric, both our approach and BAO significantly improve the baseline. And the difference between proposed approach and BAO is not obvious. We can conclude that the benefits of the proposed approach over BAO comes from the bitrate saving instead of quality enhancing.

\subsection{Qualitative Results}
\label{app:qual}
\begin{figure}[thb]
\centering
\includegraphics[width=\linewidth]{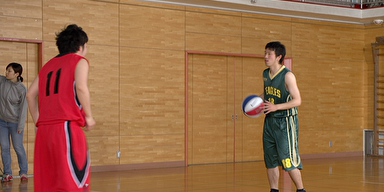}
\includegraphics[width=\linewidth]{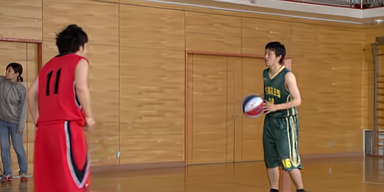}
\includegraphics[width=\linewidth]{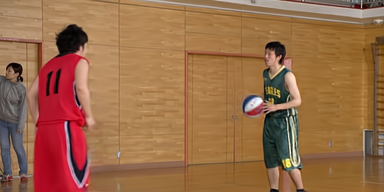}
\caption{Qualitative results using \textit{BasketballPass} of HEVC Class D. \textit{Top}. Original frame. \textit{Middle}. Baseline codec (DVC)'s reconstruction result with bpp $=0.148$ and PSNR $=33.06$dB. \textit{Bottom}. Proposed method's reconstruction result with bpp $=0.103$ and PSNR $=34.91$dB.}
\label{fig:qual1}
\end{figure}

\begin{figure}[thb]
\centering
\includegraphics[width=\linewidth]{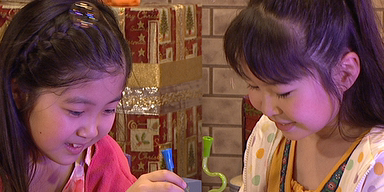}
\includegraphics[width=\linewidth]{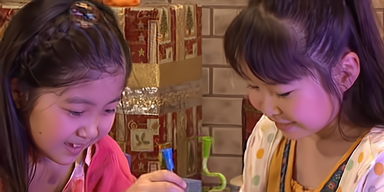}
\includegraphics[width=\linewidth]{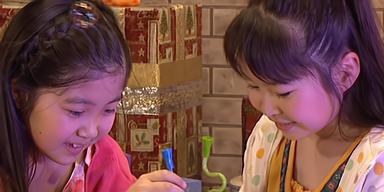}
\caption{Qualitative results using \textit{BlowingBubbles} of HEVC Class D. \textit{Top}. Original frame. \textit{Middle}. Baseline codec (DVC)'s reconstruction result with bpp $=0.206$ and PSNR $=30.71$dB. \textit{Bottom}. Proposed method's reconstruction result with bpp $=0.129$ and PSNR $=32.34$dB.}
\label{fig:qual2}
\end{figure}

\begin{figure}[thb]
\centering
\includegraphics[width=\linewidth]{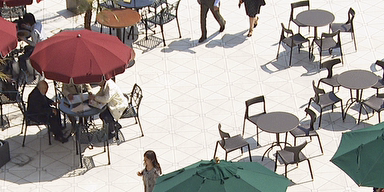}
\includegraphics[width=\linewidth]{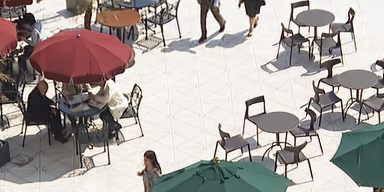}
\includegraphics[width=\linewidth]{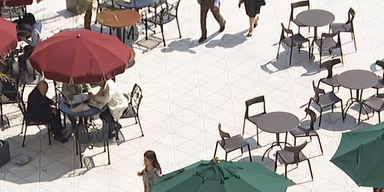}
\caption{Qualitative results using \textit{BQSquare} of HEVC Class D. \textit{Top}. Original frame. \textit{Middle}. Baseline codec (DVC)'s reconstruction result with bpp $=0.232$ and PSNR $=28.72$dB. \textit{Bottom}. Proposed method's reconstruction result with bpp $=0.128$ and PSNR $=30.87$dB.}
\label{fig:qual3}
\end{figure}

\begin{figure}[thb]
\centering
\includegraphics[width=\linewidth]{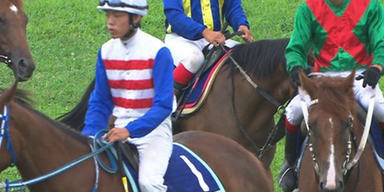}
\includegraphics[width=\linewidth]{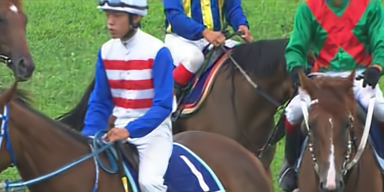}
\includegraphics[width=\linewidth]{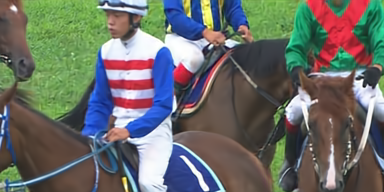}
\caption{Qualitative results using \textit{RaceHorses} of HEVC Class D. \textit{Top}. Original frame. \textit{Middle}. Baseline codec (DVC)'s reconstruction result with bpp $=0.448$ and PSNR $=30.48$dB. \textit{Bottom}. Proposed method's reconstruction result with bpp $=0.379$ and PSNR $=31.92$dB.}
\label{fig:qual4}
\end{figure}

In Fig.~\ref{fig:qual1}, Fig.~\ref{fig:qual2}, Fig.~\ref{fig:qual3} and Fig.~\ref{fig:qual4}, we present the qualitative result of our approach compared with the baseline approach. We note that compared with the reconstruction frame of baseline approach, the reconstruction frame of our proposed approach preserves significantly more details with lower bitrate, and looks much more similar to the original frame. We intentionally omit the qualitative comparison with BAO \citep{bao2022} as it is not quite informative. Specifically, from Fig.~\ref{fig:bdbrd} we can observe that the PSNR difference of BAO and our approach is very small (within $\pm 0.1$dB). And our main advantage over BAO comes from bitrate saving instead of quality improvement. Thus the qualitative difference between the proposed method and BAO is likely to fall below just noticeable difference (JND).

\subsection{More Discussion}
\label{app:disc}
Other weakness includes scalability. Our method requires jointly considering all the frame inside the GoP, which is impossible when the GoP size is large or when GoP size is unknown for live streaming tasks. Furthermore, currently the gradient ascent step number is merely chosen as an empirical sweet spot between speed and performance. A thorough grid search is desired to better understand its effect on performance.

\end{document}

%% file: math_commands.tex
\usepackage{amsmath,amsfonts,bm}

\def\eqref#1{equation~\ref{#1}}

\def\1{\bm{1}}

\DeclareMathAlphabet{\mathsfit}{\encodingdefault}{\sfdefault}{m}{sl}
\SetMathAlphabet{\mathsfit}{bold}{\encodingdefault}{\sfdefault}{bx}{n}

